\documentclass[lettersize,journal]{IEEEtran}
\usepackage{amsmath,amsfonts}
\usepackage{algorithmic}
\usepackage{algorithm}
\usepackage{array}
\usepackage[caption=false,font=normalsize,labelfont=sf,textfont=sf]{subfig}
\usepackage{textcomp}
\usepackage{stfloats}
\usepackage{url}
\usepackage{verbatim}
\usepackage{graphicx}
\usepackage{cite}
\usepackage{dsfont} 
\usepackage{algorithm}
\usepackage{algorithmic}
\usepackage{xcolor}  
\definecolor{bluejoint}{HTML}{0b0bff}  
\definecolor{orangejoint}{HTML}{fca200}  
\usepackage{pifont}
\usepackage{multirow}
\usepackage{bm}
\definecolor{myColor}{RGB}{26, 153, 230}

\usepackage{booktabs}

\newcommand{\et}{~et al.}
\definecolor{cvprblue}{rgb}{0.21,0.49,0.74}

\usepackage[pagebackref,breaklinks,colorlinks,allcolors=cvprblue]{hyperref}
\usepackage[capitalize]{cleveref}
\crefname{section}{Sec.}{Secs.}
\Crefname{section}{Section}{Sections}
\Crefname{table}{Table}{Tables}
\crefname{table}{Tab.}{Tabs.}

\usepackage{tabularx} 
\newcolumntype{A}{>{\centering\arraybackslash}X} 
\usepackage{threeparttable}   
\definecolor{bluejoint}{HTML}{0b0bff}  
\definecolor{orangejoint}{HTML}{fca200}  

\begin{document}

\title{FOAM: A General Frequency-Optimized Anti-Overlapping Framework for Overlapping Object Perception}
\author{Mingyuan Li, Tong Jia$^*$, Han Gu, Hui Lu, Hao Wang, Bowen Ma,\\ Shuyang Lin, Shiyi Guo, Shizhuo Deng, and Dongyue Chen
\thanks{
Mingyuan Li, Han Gu, Hui Lu, Hao Wang, Bowen Ma, Shuyang Lin, Shiyi Guo, Shizhuo Deng and Dongyue Chen are with the State Key Laboratory of Synthetical Automation for Process Industries, Northeastern University, Shenyang, 110819, Liaoning, China, and also with the College of Information Science and Engineering, Northeastern University, Shenyang, 110819, Liaoning, China (e-mail: 542027743@qq.com; 2400800@stu.neu.edu.cn; 2603813543@qq.com; ddsywh@yeah.net; 2010285@stu.neu.edu.cn; 2210329@stu.neu.edu.cn; guoshiyi@ise.neu.edu.cn; dengshizhuo@mail.neu.edu.cn; chendongyue@ise.neu.edu.cn).}
\thanks{Tong Jia is with the State Key Laboratory of Synthetical Automation for Process Industries, Northeastern University, Shenyang, 110819, Liaoning, China, the College of Information Science and Engineering, Northeastern University, Shenyang, 110819, Liaoning, China, and the Key Laboratory of Data Analytics and Optimization for Smart Industry, Ministry of Education, Northeastern University, Shenyang, 110819, Liaoning, China (e-mail: jiatong@ise.neu.edu.cn). \textit{(Corresponding author: Tong Jia.)}}
}
\markboth{Journal of \LaTeX\ Class Files,~Vol.~14, No.~8, August~2021}%
{Shell \MakeLowercase{\textit{et al.}}: A Sample Article Using IEEEtran.cls for IEEE Journals}


\maketitle


\begin{abstract}

Overlapping object perception aims to decouple the randomly overlapping foreground-background features, extracting foreground features while suppressing background features, which holds significant application value in fields such as security screening and medical auxiliary diagnosis. 
Despite some research efforts to tackle the challenge of overlapping object perception, most solutions are confined to the spatial domain.
Through frequency domain analysis, we observe that the degradation of contours and textures due to the overlapping phenomenon can be intuitively reflected in the magnitude spectrum. Based on this observation, we propose a general Frequency-Optimized Anti-Overlapping Framework (FOAM) to assist the model in extracting more texture and contour information, thereby enhancing the ability for anti-overlapping object perception.
Specifically, we design the Frequency Spatial Transformer Block (FSTB), which can simultaneously extract features from both the frequency and spatial domains, helping the network capture more texture features from the foreground.
In addition, we introduce the Hierarchical De-Corrupting (HDC) mechanism, which aligns adjacent features in the separately constructed base branch and corruption branch using a specially designed consistent loss during the training phase. This mechanism suppresses the response to irrelevant background features of FSTBs, thereby improving the perception of foreground contour.
We conduct extensive experiments to validate the effectiveness and generalization of the proposed FOAM, which further improves the accuracy of state-of-the-art models on four datasets, specifically for the three overlapping object perception tasks: Prohibited Item Detection, Prohibited Item Segmentation, and Pneumonia Detection. The code will be open source once the paper is accepted.

\end{abstract}


\section{Introduction}
Overlapping object perception, a fundamental task within the realm of computer vision, involves tasks like prohibited item detection, prohibited item segmentation, and pneumonia detection. These tasks aim to decouple the randomly overlapping foreground-background features and extract foreground features while suppressing background features, holding significant application value in fields such as security inspection and medical auxiliary diagnosis.
\begin{figure}
    \centering
    \includegraphics[width=1\linewidth]{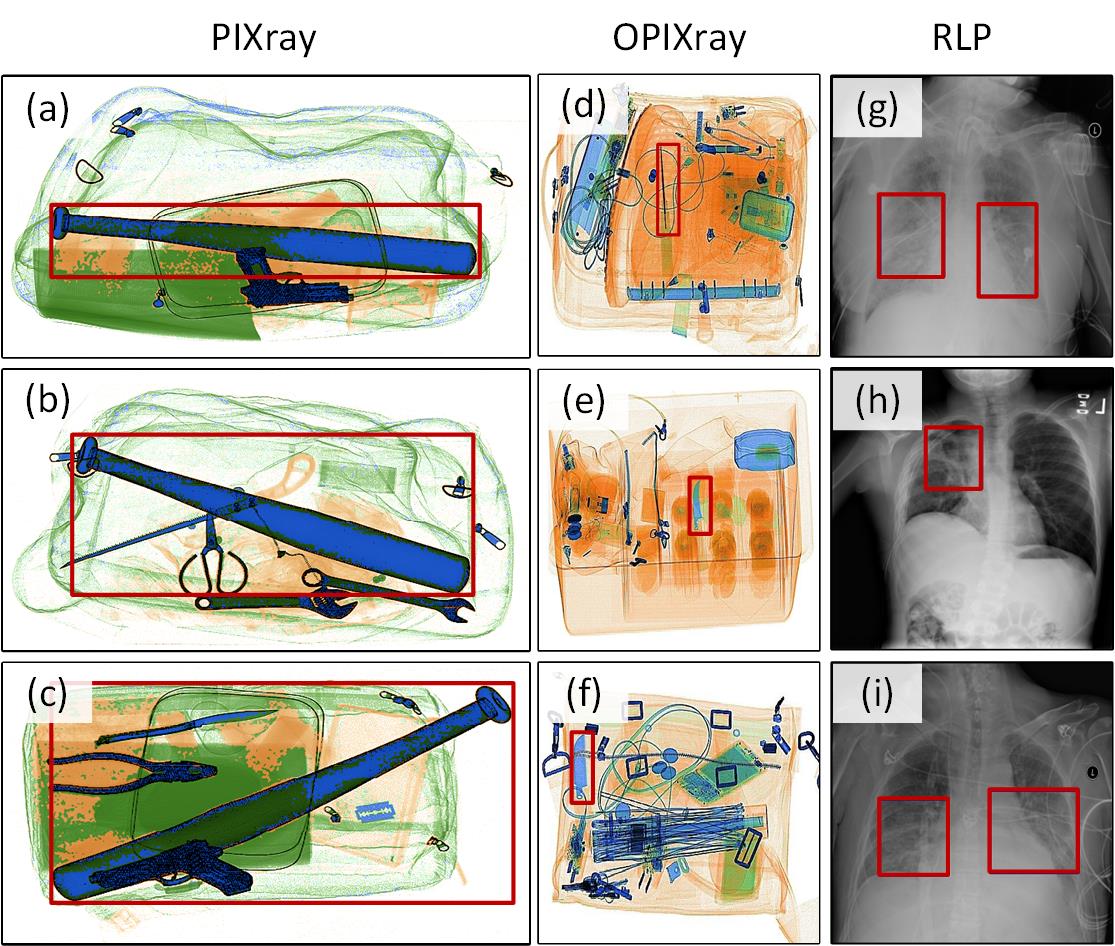}
    \caption{
    The overlapping phenomenon causes the contours and textures of foreground objects to be randomly degraded in different background scenarios. (a)-(c) show the baseball ``Bat'' in the PIXray~\cite{PIXray} dataset, (d)-(f) display the ``Straight Knife'' in the OPIXray~\cite{OPIXray} dataset, and (g)-(i) depict ``Lung Opacity'' in the RLP subset of the RSNA~\cite{RSNA_Pneumonia_Detection_Challenge2018} dataset.
    }
    \label{fig:samples of overlapping}
\end{figure}

Fig.~\ref{fig:samples of overlapping} illustrates typical overlapping phenomena in the PIXray, OPIXray, and RLP datasets. The contours and textures of foreground objects, such as the metal baseball ``Bat'' and ``Straight Knife'', in subfigures (a)-(c) and (e)-(f), are degraded to varying degrees by randomly occurring backgrounds. Therefore, in real-world security inspection scenarios, even the most advanced general vision models struggle to perform accurate object detection or instance segmentation of meticulously concealed prohibited items in X-ray images. Similarly, in subfigures (g)-(i), the imaging texture and contour of ``Lung Opacity'' are unclear, making them highly susceptible to interference from background elements~\cite{gambato2023chest}, such as potential EKG leads, external tubes, artifacts, overlapping devices, bones, and healthy tissues. Therefore, in real-world medical auxiliary diagnosis scenarios, general vision models face significant challenges in performing accurate object detection on diverse pathological tissues.

Recently, an increasing number of studies have proposed advanced deep learning techniques to specifically address this task, especially after the introduction of several large-scale pseudo-colored X-ray datasets.
Specifically, methods such as GADet~\cite{GADet}, AO-DETR~\cite{AO-DETR}, Xdet~\cite{Xdet}, and CLCXray~\cite{CLCXray} introduce label assignment strategies to ensure that models consistently focus on high-quality foreground objects during training. OPIXray~\cite{OPIXray}, SIXray~\cite{SIXray}, and PID-YOLOX~\cite{PID-YOLOX} leverage attention mechanisms to assist models in decoupling and extracting foreground features from overlapping foreground and background. Additionally, PIXray~\cite{PIXray} and AO-DETR~\cite{AO-DETR} adopt multistage regression approaches to perceive blurred contours.
However, the aforementioned methods rely on spatial domain information to perceive contraband, making the models susceptible to interference and deception from unknown texture and contour in the background~\cite{FDRNet}. 

A promising novel approach is to leverage frequency domain learning to enhance the contour and texture details of the foreground, complement spatial domain information, and thereby improve the model's ability to perceive overlapping objects.
Specifically, FAPID~\cite{FAPID} truncates the frequency domain information obtained from the Fast Fourier Transform (FFT)~\cite{FFT} using a fixed high-pass filter, which serves as contour and texture cues to correct spatial domain features. FDTNet~\cite{FDTNet} uses a CBAM-like attention mechanism to refine the local frequency domain information obtained from the SRM~\cite{SRM} filter, aiming to adaptively extract informative frequency information to complement the spatial domain information. However, the implementation of a fixed high-pass filter in FAPID results in the exclusion of valuable low-frequency information, whereas the use of the SRM filter in FDTNet~\cite{FDTNet} is limited in its global perception capabilities when compared to FFT. 
Therefore, although these two approaches validate and demonstrate the effectiveness and research value of frequency domain learning for overlapping object perception tasks, their understanding and development are limited, leaving significant room for improvement.

\begin{figure}
    \centering
    \includegraphics[width=1\linewidth]{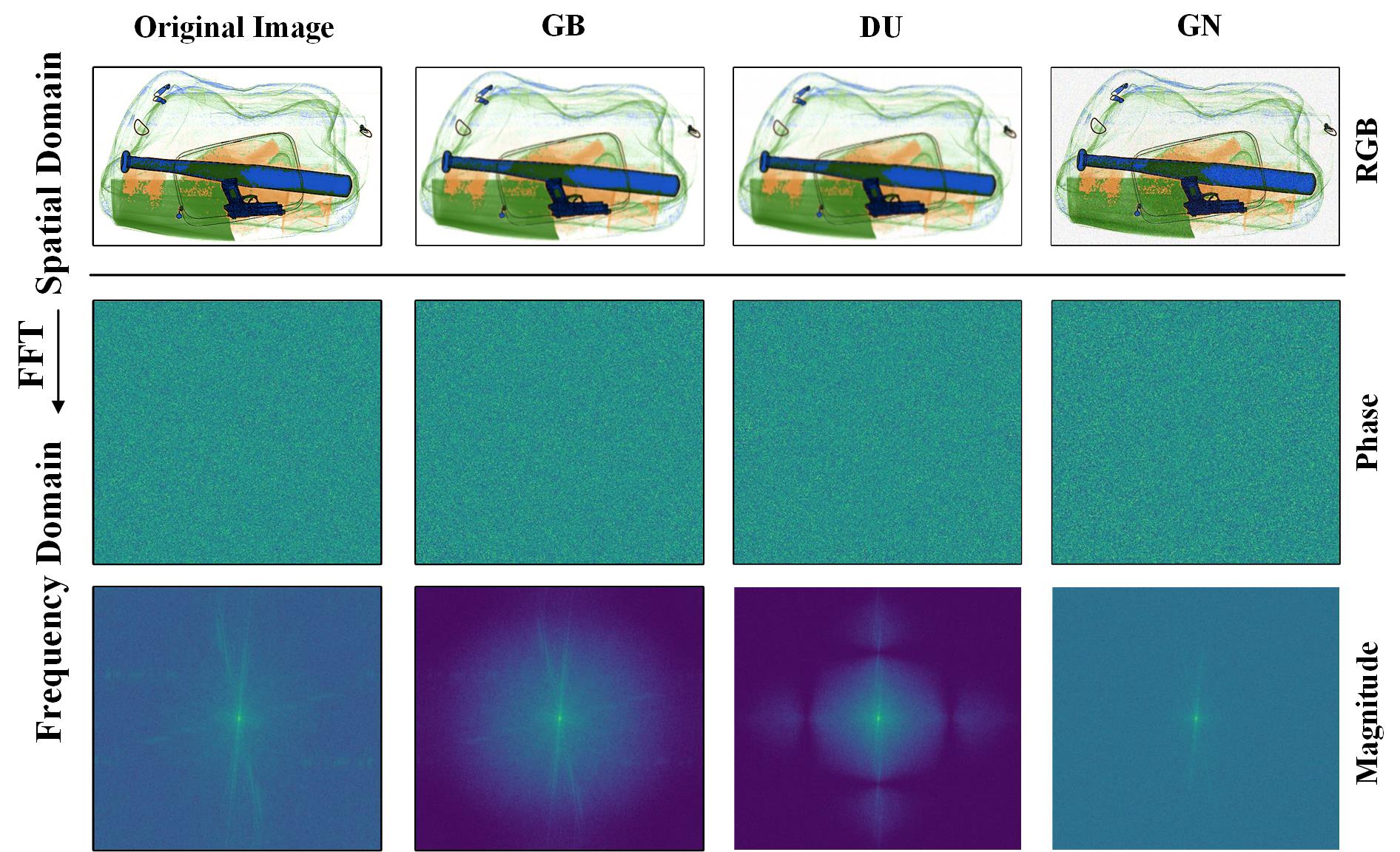}
    \caption{X-ray images analysis in spatial and frequency domains. GB, DU, and GN represent Gaussian Blurring, Downsampling and Upsampling, and Gaussian Noise corruption strategies, respectively.}
    \label{fig:frequency analysis}
\end{figure}

We conduct an in-depth analysis of the characteristic representations of the magnitude spectrum and phase spectrum derived from the frequency domain information obtained through FFT. As shown in Fig.~\ref{fig:frequency analysis}, after applying classic corruption methods such as Gaussian Blurring (GB), Downsampling and Upsampling (DU), and Gaussian Noise (GN), the contours and textures information of the images are compromised. 
Obviously, neither the spatial domain image nor the phase spectrum in the frequency domain, which is adept at capturing object shapes and structures, shows significant changes. In contrast, the magnitude spectrum is more sensitive to changes in texture and contour information, revealing underlying patterns in the image that are not easily observed from the raw pixel values. For example, in the magnitude spectrum, it can be observed that GB and DU primarily remove mid- to high-frequency fine details from the image, while GN injects mid- to high-frequency noise. Therefore, we believe that the magnitude spectrum provides informative information for frequency domain learning for overlapping object perception tasks, facilitating the decoupling of foreground and background information, and the experimental results in~\cref{sec:Ablation study of SDCA and FDBA} corroborate our theory.



 Based on this observation, we propose a general Frequency-Optimized Anti-Overlapping Framework (FOAM) for accurate overlapping object perception, which combines global frequency features with local spatial features to capture texture and contour information, thereby enhancing feature discriminability.
 
To implement FOAM, we first design a fundamental building block named Frequency Spatial Transformer Block (FSTB), which simultaneously extracts features from both the spatial and frequency domains and helps the network extract more texture features from the foreground. It consists of three components: the Frequency Domain Bands Self-Attention (FDBA) mechanism, the Spatial Domain Channel Self-Attention (SDCA) module, and the Frequency Spatial Feed-forward Network (FSFN). FDBA mechanism leverages the global dependency relationship among different frequency bands to reconstruct their proportions while keeping the phase unchanged, thereby correcting the texture and contour information perceived in the spatial domain. The lightweight SDCA module optimizes the local spatial details representation of objects. FSFN is responsible for the integration and optimization of both spatial and frequency domain information.
Then, as a fundamental unit, FSTB is utilized to iteratively optimize the features of the backbone for $N$ iterations, and obtains the basic feature set. This branch is referred to as the base branch and is enabled during both training and inference.

Furthermore, we propose a Hierarchical De-Corrupting (HDC) mechanism, which establishes a corruption branch that is enabled only during the training phase. We apply corruption strategies to the original image to simulate the blurring of textures and contours of the current foreground object under more severe overlapping phenomena, as well as the introduction of background noise, resulting in a low-quality corrupted image. Subsequently, we optimize the features through a shared-weight backbone and FSTBs to obtain the corrupted feature set. Finally, we employ a consistent loss to align the features from the corruption branch with those from the base branch, guiding the FSTBs to suppress the response to irrelevant background features and adapting to the anti-overlapping perception task.

We utilize FOAM across four datasets to further improve the accuracy of state-of-the-art models on three overlapping object perception tasks: Prohibited Item Detection, Prohibited Item Segmentation, and Pneumonia Detection.

Our main contributions are summarized as follows:
\begin{enumerate}
\item{We propose the Frequency-Optimized Anti-Overlapping Framework (FOAM), which leverages both frequency domain and spatial domain cues to help models capture more texture and contour under the negative impact of overlapping scenes for object perception. This architecture is designed to be compatible with most CNN-based and Transformer-based object detection and instance segmentation models.}
\item{
To improve the comprehensive understanding capability of texture features, we design the Frequency Spatial Transformer Block (FSTB) to simultaneously extract foreground cues from both the frequency domain and the spatial domain.}
\item{
To improve the perception ability for the foreground contour of networks, we propose the Hierarchical De-Corrupting (HDC) mechanism, which utilizes features from the base branch to supervise the features from the corruption branch during the training phase, suppressing the response to irrelevant background features of FSTBs.

}
\end{enumerate}
\section{Related Work}
\subsection{Frequency Domain Learning}
Frequency domain information, distinct from spatial domain information, represents a unique form of high-order information with global feature representation, offering a distinctive perspective for image processing and understanding. Therefore, frequency domain learning has often been utilized for analysis and applications in the fields of image compression and super-resolution~\cite{fritsche2019frequency,vandewalle2006frequency,grgic2001performance,velisavljevic2007space}. 
Recently, some works~\cite{Xu_2020_CVPR,zhou2024general,sun2025frequency,ding2023enhanced,liu2024edge,zhou2022spatial,cui2023image} on frequency domain learning have made progress in remote sensing and camouflaged object detection, sparking a wave of exploration among visual perception researchers. 
Specifically, Xu\et\cite{Xu_2020_CVPR} builds upon the SE-block~\cite{SENet} and proposes a learning-based dynamic channel selection method to identify trivial frequency components for static removal during inference, which is the first work to explore frequency domain learning in object detection and instance segmentation.
FcaNet~\cite{Fcanet} proposes to leverage frequency domain learning to address the information loss problem in channel attention mechanisms. SPANet~\cite{SPANet} handles the balancing problem of high- and low-frequency components in visual features.
However, the aforementioned work did not explore the interaction between RGB images and frequency domain cues. Zhong\et \cite{zhong2022detecting} applies the Discrete Cosine Transform (DCT)~\cite{DCT} to every 8×8 patch to extract frequency domain clues and uses a multi-head attention mechanism to combine frequency domain information with RGB domain information. The frequency domain information obtained through DCT consists of real-valued data, lacking the representation of the phase spectrum feature that is critical for capturing object position and structural details. Moreover, the approach of dividing features into patches before transformation causes the frequency domain representation to lose the advantage of global perception.
FSEL~\cite{FSEL} further employs the Fast Fourier Transform (FFT) to extract global frequency domain clues and integrates frequency domain and spatial domain information using a variant of the self-attention mechanism. However, this work lacks the independent design based on the distinct characteristics of magnitude and phase.
 In the domain of prohibited item detection, to the best of my knowledge, there is only two relevant works.  FAPID~\cite{FAPID} truncates the frequency domain information obtained from the Fast Fourier Transform (FFT)~\cite{FFT} using a fixed high-pass filter, which serves as contour and texture cues to correct spatial domain features, whereas completely ignoring the low-frequency information. FDTNet~\cite{FDTNet} uses an SRM filter to provide frequency domain information, but its filtering approach is essentially a set of fixed large kernel convolutions, which lack the global perspective compared to the frequency domain information obtained from Fourier transforms. 

In this paper, we delve into the characteristics of the frequency domain signals obtained through FFT transformation, decoupling the learning of magnitude and phase, and combine frequency domain clues with RGB domain information to extract the contour and texture details of foreground objects.
\subsection{Attention Mechanism in Computer Vision}
The main goal of the attention mechanism is to help the model mimic the human visual system's ability to focus on foreground information in images rather than irrelevant background. This mechanism can typically be divided into CNN-based attention mechanisms and Transformer-based self-attention mechanisms.

For CNN-based attention mechanisms, SENet~\cite{SENet} proposes the most well-known channel attention mechanism, which compresses features into a vector using global average pooling, and then applies a fully connected layer to weight the features of each layer. GE~\cite{GE} employs spatial attention to better exploit the feature context. Building upon these works, models like CBAM~\cite{CBAM}, CA~\cite{CA}, BAM~\cite{BAM}, DAN~\cite{DAN}, and PID-YOLOX~\cite{PID-YOLOX} explore the integration of spatial and channel attention.
CBAM~\cite{CBAM} argues that global average pooling leads to information loss, prompting the introduction of global max pooling, which achieves superior performance. Inspired by this, GSoP~\cite{GSoP} and SRM~\cite{SRM} further incorporate second-order pooling and global standard deviation pooling. SkNet~\cite{SkNet} and ResNeSt~\cite{ResNeSt} propose selective channel aggregation and attention mechanisms.

For the Transformer-based self-attention mechanism, the Transformer~\cite{Attention_is_all_you_need} was originally designed by Vaswani et al. for NLP tasks, relying solely on attention mechanisms and dispensing with recurrence and convolutions entirely. The Transformer has the ability to model global semantics and long-range dependencies, and its ideas have inspired many works in computer vision, including classification models like ViT~\cite{ViT} and PVT~\cite{PVT}, object detection models like DETR~\cite{DETR} and Deformable-DETR~\cite{Deformable-DETR}, and instance segmentation models like CondInst~\cite{CondInst}, Cascade-Mask-R-CNN~\cite{Cascade_Mask_R-CNN}, MaskFormer~\cite{MaskFormer}, and Mask2Former~\cite{Mask2Former}. For example, ViT~\cite{ViT} divides images into independent patches to reduce the cost of capturing long-range relationships. The Swin Transformer~\cite{Swin-Transformer} further enhances model efficiency through a shift operation. Additionally, other works, such as EfficientViT~\cite{EfficientViT} and PVTv2~\cite{PVTv2}, have also achieved good performance. 

However, these methods focus on global modeling of spatial domain features, neglecting the powerful representational capability of frequency domain information for textures and contours. Therefore, we aim to propose a transformer-based attention mechanism for integrating and refining both spatial and frequency domain features to enhance the model's ability to perceive informative foreground elements in X-ray images.
\subsection{Prohibited Item Perception}
Following the introduction of the first pseudo color large-scale X-ray image dataset, SIXray~\cite{SIXray}, a multitude of modern pseudo color X-ray prohibited perception datasets has been developed. These datasets are specifically designed for various tasks, including classification with SIXray, object detection using OPIXray, PIXray-det~\cite{PIXray}, HIXray~\cite{HIXray}, PIDray-det~\cite{PIDray}, DvXray~\cite{DvXray}, and CLCXray~\cite{CLCXray}, as well as segmentation with PIXray-seg~\cite{PIXray} and PIDray-seg~\cite{PIDray}. Given the significant practical application value of object detection tasks, the majority of contemporary prohibited item perception methodologies predominantly focus on the development and enhancement of prohibited item detectors.
Most of them are optimized based on traditional object detectors to cater to the unique imaging characteristics of X-ray images. Specifically, SIXray introduces a feature pyramid network FPN-like approach named class-balanced hierarchical refinement, which supervises lower-level features with higher-level features, thereby enhancing the focus on foreground information. OPIXray~\cite{OPIXray} and OVXD~\cite{OVXD} propose the DOAM module and bottleneck-like adapter, respectively, which emphasize foreground materials and contour information. GADet~\cite{GADet} and Xdet~\cite{Xdet} present the IAA and HSS labeling strategies, respectively, to improve foreground perception accuracy by alleviating the issue of class imbalance between foreground and background categories. AO-DETR~\cite{AO-DETR} is the first to introduce a DETR-like architecture, DINO~\cite{DINO}, in the field of prohibited detection, proposing the CSA strategy to train category-specific content queries that are specifically responsible for perceiving particular categories of contraband. Furthermore, MMCL~\cite{MMCL} and CSPCL~\cite{CSPCL} propose plug-and-play contrastive learning strategies to address the issue of distribution imbalance of category-specific content queries in Deformable-DETR-like models.
However, the aforementioned methods are all limited to spatial domain feature perception and fail to utilize the representational power of frequency domain information regarding contours and textures. FDTNet and FAPID employ convolutional or self-attention mechanisms to extract frequency domain information, thereby acquiring contour and texture information from foreground objects. However, their frequency domain information is derived from the SRM filter or high-pass filter, which incurs more feature loss compared to FFT transformation, resulting in them being only locally optimal solutions.
After decoupling it into magnitude and phase spectra with distinct characteristics, we design a frequency domain attention mechanism in an attempt to identify the optimal strategy for frequency domain information extraction.
\section{Methodology}
\begin{figure*}
    \centering
    \includegraphics[width=1\linewidth]{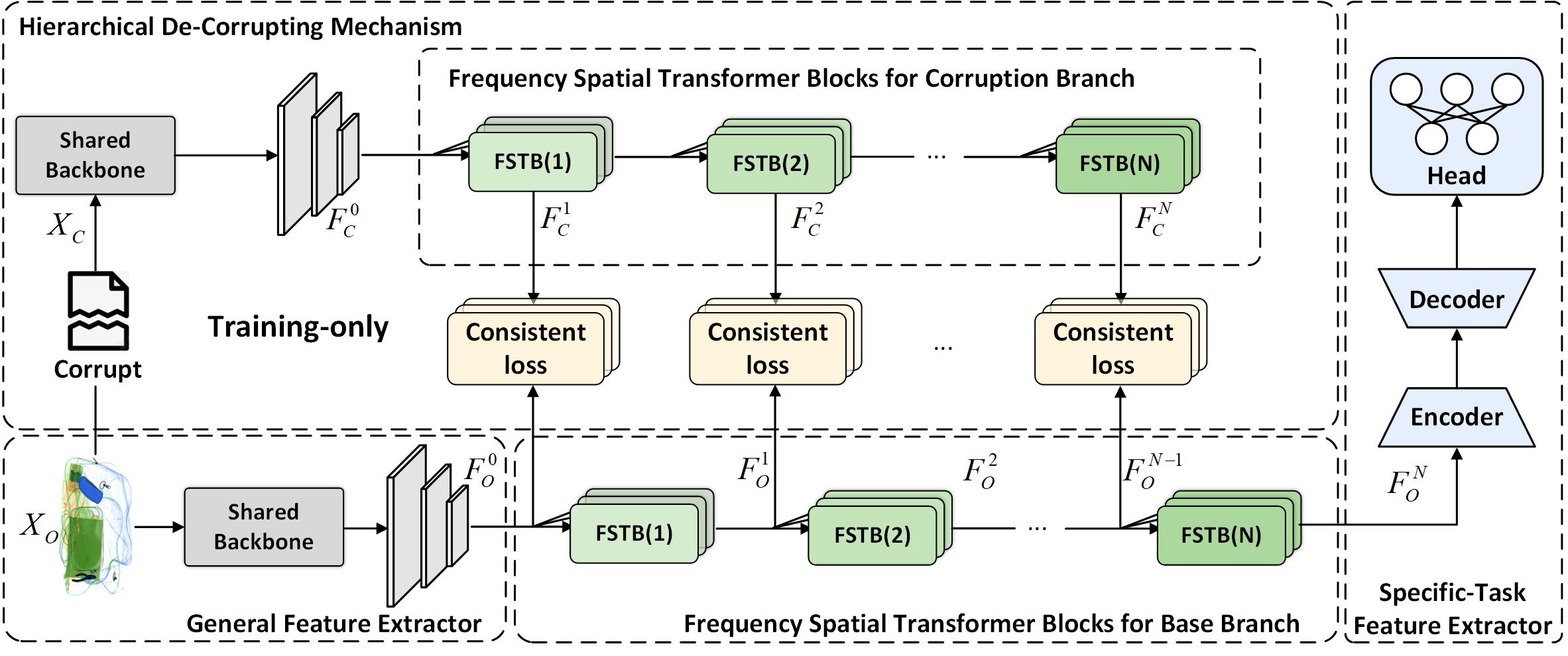}
    \caption{Overall architecture of our proposed FOAM.
    The core module is FSTB, which is designed for joint learning in both the frequency and spatial domains, enabling the backbone network to effectively perceive and extract texture features, including both foreground and background. The HDC mechanism first corrupts the original image $X_O$ to obtain the corrupted image $X_C$. After feature extraction through a shared-parameter extractor, cascaded FSTBs are then applied to generate the base branch and the corruption branch. Finally, a consistent loss is employed to align adjacent features from the two branches, suppressing the FSTB's response to irrelevant background features and clarifying the edges of foreground objects.
    }
    \label{fig:overall architecture}
\end{figure*}

In this section, we first introduce the principles and properties of image Fourier transformation. We then present the overall structure of the proposed Frequency-Optimized Anti-Overlapping Framework (FOAM), as depicted in~\cref{fig:overall architecture}. Subsequently, we describe the core module, the Frequency Spatial Transformer Block (FSTB), which internally includes the Frequency Domain Bands Self-Attention (FDBA) mechanism, the Spatial Domain Channel Self-Attention (SDCA) module, and the Frequency Spatial Feed-forward Network (FSFN). Finally, we explain how the Hierarchical De-Corrupting (HDC) mechanism utilizes the proposed consistent loss to deeply activate the anti-overlapping perception capability of the FSTB.
\subsection{Image Fourier Transformation}
\label{sec: Image Fourier Transformation}
2D Discrete Fourier Transform (2D DFT) is a widely used method for analyzing the frequency content of images. 
Compared to the 2D Discrete Cosine Transform (2D DCT), it additionally provides phase information, offering a comprehensive representation and description about location and structure of the objects~\cite{Digital_Image_Processing_3rd_Edition}. In contrast to the 2D Discrete Wavelet Transform (2D DWT)~\cite{Digital_Image_Processing_3rd_Edition}, it dynamically transforms low and high-frequency elements and does not need predefined kernels to separate frequency components~\cite{SFINet}, thereby providing more informative clues about the detection of foreground objects for the model. For multi-channel image signals, the Fourier transform is typically applied to each channel individually. To simplify, we omit the channel notation in the subsequent equations. Let $X\in \mathds{R}^{C\times H\times W}$ represent the image, and the 2D DFT maps it to the complex components in the Fourier space $F(u,v)$, which can be expressed as:

\begin{equation}
\label{fourier}
F(u,v)=\sum_{h=0}^{H-1}\sum_{w=0}^{W-1}X(h,w)e^{-j2\pi(\frac{h}{H}u+\frac{w}{W}v)},
\end{equation}
where $F(u, v) \in \mathbb{C}^{H \times W}$. $u\in \{0,1,...,H-1\}$ and $v\in \{0,1,...,W-1\}$ represent the vertical and horizontal frequency indices, respectively. The 2D Inverse Discrete Fourier Transform (2D IDFT) is denoted as:
\begin{equation}
\label{invert fourier}
X(h,w)=\frac{1}{HW}\sum_{u=0}^{H-1}\sum_{v=0}^{W-1}F(u,v)e^{j2\pi(\frac{h}{H}u+\frac{w}{W}v)}.
\end{equation}
In our work, we employ efficient and equivalent Fast Fourier Transform (FFT) and Invert Fast Fourier Transform (IFFT)~\cite{FFT} to replace the 2D DFT and its inverse transform, processing each image channel individually, as followed in~\cite{SFINet,FSEL}.

The magnitude component $M(u, v)$ and phase component $P(u, v)$ are defined as:

\begin{equation}
\label{magnitude}
M(u,v)=\sqrt{R^2(u,v)+I^2(u,v)},
\end{equation}
\begin{equation}
\label{phase}
P(u,v)=\text{arctan}(\frac{I(u,v)}{R(u,v)}),
\end{equation}
\begin{equation}
\label{R}
R(u,v)=(F(u,v)+conj(F(u,v)))/2,
\end{equation}
\begin{equation}
\label{I}
I(u,v)=(F(u,v)-conj(F(u,v)))/2j.
\end{equation}
Here, $R(u,v)$ and $I(u,v)$ represent the real and imaginary parts of $F(u,v)$ respectively, and $conj(\cdot)$ denotes the conjugate complex number operator.

As shown in~\cref{fig:frequency analysis}, the frequency spectrum and the phase spectrum emphasize different aspects of image representation. The former is adept at capturing texture and contour information, while the latter excels in capturing shape and structural information~\cite{Digital_Image_Processing_3rd_Edition}. 
In~\cref{sec:FSTB}, we design a joint learning method that combines the frequency and spatial domains, leveraging their complementary information based on the aforementioned characteristics.

\subsection{Frequency-Optimized Anti-Overlapping Framework}

As shown in~\cref{fig:overall architecture}, the architecture of FOAM differs between the training and inference phases. In the inference phase, the network structure only requires the insertion of the base branch, constructed using the FSTB proposed in~\cref{sec:FSTB}, into the general object perception model. This allows for fine-tuning the features extracted by the backbone network, thereby aiding the Specific-Task Feature Extractor in better handling the overlapping object perception task. In the training phase, the HDC mechanism is additionally introduced, where the corruption branch is constructed to supervise the features of the base branch. This process guides the FSTB to reduce its response to background features, further enhancing the perception ability for the foreground contour.

\subsubsection{Base Branch}
The base branch is enabled during both training and inference, realized by a set of FSTB modules inserted into the original model, whose details will be presented in~\cref{sec:FSTB}. Specifically, given the original image $X_O$, the network first applies a standard backbone, \textit{i.e.}, ResNet, ResNeXt, and Swin Transformer, to obtain the initial multi-scale features $F_O^0=\{F_{O,l}^0\}_{l=0}^L$ for the base branch, where $L$ are the stage number of the backbone. Then, $N$ independent FSTBs are cascaded together to adaptively perform entanglement learning on $F_O^0$ from both the spatial and frequency domains, resulting in a discriminative feature set $F_O=\{F_O^1, F_O^2, ..., F_O^N\}$. Among them, the feature enhanced by $n$ FSTB modules is expressed as follows:
\begin{equation}
\label{equation FSTB base}
F_O^n=(FSTB^n \circ FSTB^{n-1} \circ ...\circ FSTB^1)(F_O^0),
\end{equation}
where $\circ$ is the composition operator, and $FSTB^n$ represents the $n$-th FSTB operator.
Finally, the prediction results are obtained through a specific-task feature extractor, consisting of the encoder, decoder, and head for specific-task. 
\subsubsection{Hierarchical De-Corrupting Mechanism}
HDC mechanism is enabled only during the training phase. Corruption operation, such as Gaussian Blurring (GB), Downsampling and Upsampling (DU), and Gaussian Noise (GN), is applied to the original image to obtain a corrupted image $X_C$, which further disrupts the textures and contours of overlapped target objects to simulate more severe overlapping scenes.
Similarly to the base branch, the initial multi-scale features $F_C^0$ for the corruption branch is obtained through a backbone with shared weights, followed by $N$ FSTB operations, whose parameters are consistent with those of the $N$ FSTBs in the base flow, resulting in the corruption version of the discriminative feature set $F_C=\{F_C^1, F_C^2, ..., F_C^N\}$.
Theoretically, since the corruption operations disrupt the texture and outline details, the effective features in the corrupted feature $F_C^n$ at the $n$-th stage are fewer than those in the base branch $F_O^n$. Ideally, the FSTB is designed to extract and enhance relevant information. Therefore, the quality of discriminative information in $F_C^{n+1}$, which undergoes an additional FSTB operation, is superior to that in $F_C^{n}$.
We propose a Type I consistent loss based on KL divergence, which employs the strategy of supervising $F_C^{n+1}$ with $F_O^n$ to achieve fine-grained alignment of the two multi-scale feature sets $F_C$ and $F_O$, thereby directing FSTBs to enhance the de-corruption capability in extracting and reinforcing contours and textures. The process is as follows:
\begin{equation}
\label{equation KL loss}
L_C^I(F_O,F_C)=-\sum_{n=1}^{N}\sum_{l\in \mathcal{L}}\sum_{i=1}^{HW}\hat{F}_{O,l}^{n-1}(i)\text{log}(\frac{\hat{F}_{O,l}^{n-1}(i)}{\hat{F}_{C,l}^n(i)}),
\end{equation}
\begin{equation}
\label{equation KL loss 2}
\hat{F}_{O,l}^{n-1}(i)=\frac{\text{exp}(F_{O,l}^{n-1}(i))}{\sum_{i=1}^{HW}\text{exp}(F_{O,l}^n(i))},
\end{equation}
\begin{equation}
\label{equation KL loss 3}
\hat{F}_C^n(i)=\frac{\text{exp}(F_{C,l}^n(i))}{\sum_{i=1}^{HW}\text{exp}(F_{C,l}^n(i))},
\end{equation}
where $\mathcal{L}$ is the target layer set involved in fine-grained alignment. Note that $F_O^n$ is multi-scale features with $L$ scales, where the higher-level features typically extract abstract global information, excelling in representing semantic characteristics and contextual relationships. In contrast, the lower-level features often capture local information but may contain redundant information and noise. Therefore, it is necessary to select the appropriate layer set $\mathcal{L}$ for feature alignment in order to achieve an optimal balance. Related ablation experiment results and analysis are shown in~\cref{sec:Ablation study for target layer set}. 

The exact form of the consistent loss is not crucial. We also propose an alternative variant of the consistent loss that possesses similar properties and yields comparable results. This variant is designed based on the MSE loss and is referred to as the Type II consistent loss, as expressed in the following formula:
\begin{equation}
\label{equation MSE loss}
L_C^{II}(F_O,F_C)=\frac{1}{HW}\sum_{n=1}^{N}\sum_{l\in L}\sum_{i=1}^{HW}(F_{O,l}^{n-1}(i)-F_{C,l}^n(i))^2.
\end{equation}



The experimental results comparing the consistent loss of Type II and Type I are presented in~\cref{sec:Ablation study for target layer set}. Both of them achieve the global optimal value when $\{\{F_{O,l}^n\}_{n=0}^{N-1}\}_{l\in\mathcal{L}}$ and $\{\{F_{C,l}^n\}_{n=1}^{N}\}_{l\in\mathcal{L}}$ are perfectly aligned, indicating that FSTBs has obtained ideal de-corruption capabilities by reducing the model’s response to background noise and blur caused by overlapping phenomena. According to Theorem I, this suppression ability of background features ultimately manifests as an enhancement in contour perception, which is also corroborated by the feature map visualizations in~\cref{sec:Feature maps}.

\textbf{Theorem I:} \textit{Assume that in an overlapping scene, a homogeneous foreground overlaps with a homogeneous background. Let positive numbers $f$ and $b$ represent the response values of the neural network to the foreground and background, respectively. In the ideal linear case, the response value of the foreground region is $f+b$. The contrast at the foreground contour is defined as: $\frac{f+b}{b}$. When the model’s response to the background decreases by $c$, where $ c<b$, the following inequality holds:}
\begin{equation}
\label{equation HDC}
\frac{f+b-c}{b-c}>\frac{f+b}{b}.
\end{equation}

\begin{figure*}
    \centering
    \includegraphics[width=1\linewidth]{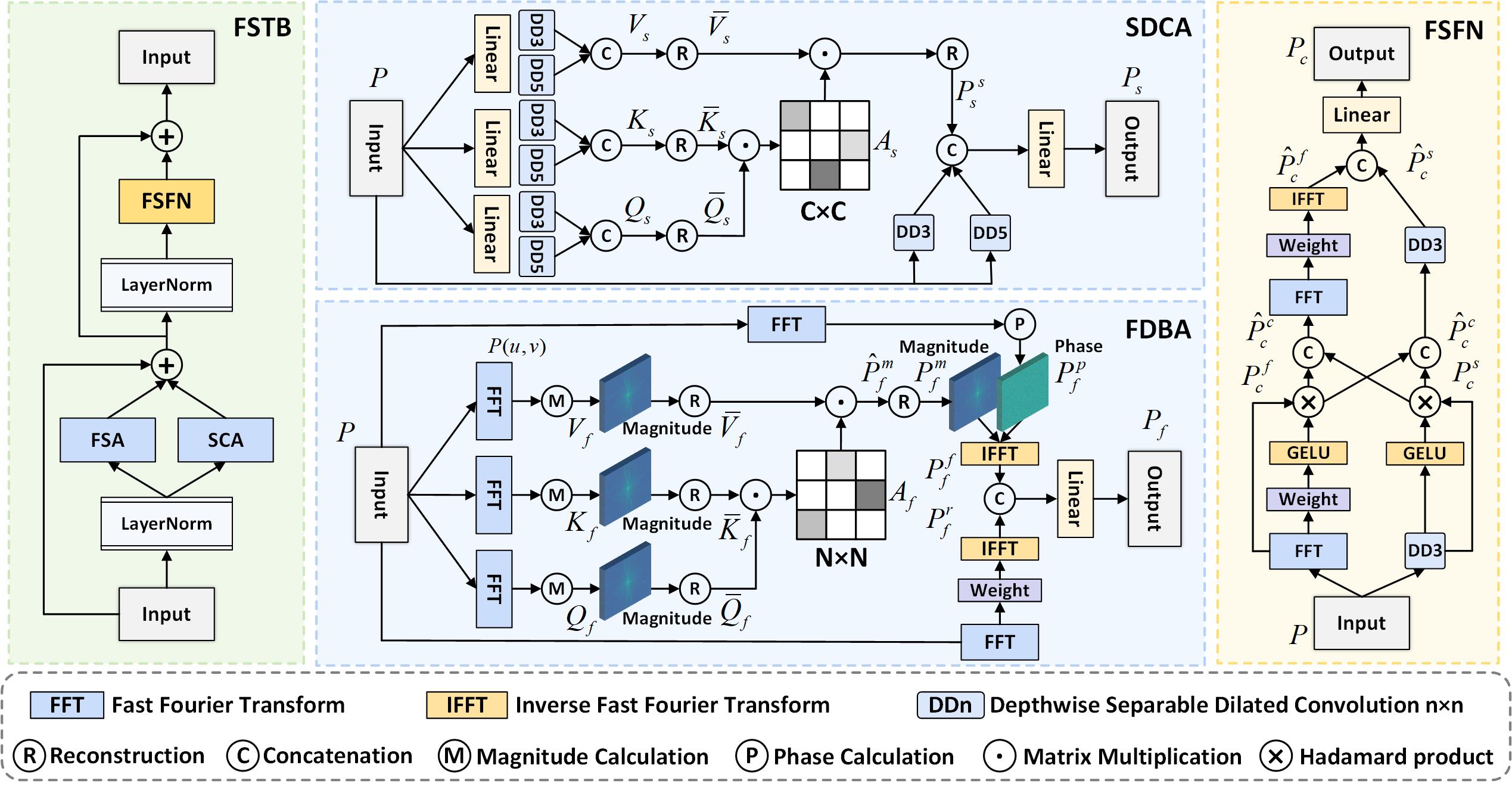}
    \caption{The detailed flowchart of the proposed core building module FSTB, consisting of three components: Spatial Domain Channel Self-Attention (SDCA), Frequency Domain Bands Self-Attention (FDBA), and Frequency Spatial Feed-forward Network (FSFN).}
    \label{fig:overall architecture 2}
\end{figure*}
\subsection{Frequency Spatial Transformer Block}\label{sec:FSTB}
Unlike previous methods~\cite{FAPID} and~\cite{FDTNet}, which use a constant high-pass filter and a fixed SRM filter, respectively, to attempt to filter low-frequency features while preserving texture and contour information, or methods~\cite{FPNet,Swin-Transformer} only model long-range dependencies based on local features in the spatial domain.
Our FSTB integrates information from both the frequency and spatial domains simultaneously. It utilizes different adaptive learning approaches for the two domain features, allowing for a dynamic and targeted understanding and integration of information such as color, texture, edges, spectral characteristics, magnitude, and energy. This entanglement learning approach facilitates the learning of discriminative foreground features from coupled features. 
As depicted in~\cref{fig:overall architecture 2}, the proposed FSTB consists of three key components: Spatial Domain Channel Self-Attention (SDCA), Frequency Domain Bands Self-Attention (FDBA), and Frequency Spatial Feed-forward Network (FSFN).
First, SDCA employs a variant of the self-attention mechanism to capture local information in the spatial domain. Second, FDBA combines the attention mechanism with Fourier transformations to extract global representations from the frequency domain. Third, FSFN enhances the information flow between the frequency and spatial domains, facilitating the learning of complementary representations
\subsubsection{SDCA}
In the spatial domain, we design a channel-oriented self-attention variant that can capture long-range interdependencies among channels. This mechanism adaptively recalibrates the responses of each channel's features, thereby enhancing the latent foreground information perceived in the spatial domain.
Specifically, as illustrated in~\cref{fig:overall architecture 2}, given the input feature $P\in \mathds{R}^{C\times H\times W}$, we obtain the positional embedding through a $1\times1$ convolution, following~\cite{Attention_is_all_you_need}. Then, two depthwise separable dilated convolutions are used to generate the query the $query$ $Q_s$, $key$ $K_s$, and $value$ $V_s$. For example, the calculation process of $Q_s$ is as follows:
\begin{equation}
\label{SDCA}
Q_s=\text{Cat}(\text{DD}_3^\frac{C}{2}(\text{Conv}_1^C(P)),\text{DD}_5^\frac{C}{2}(\text{Conv}_1^C(P)))
\end{equation}
Here, $Q_s=\in \mathds{R}^{C\times H\times W}$ with $C=256$. 
$\text{DD}_k^C$ and $\text{Conv}_k^C$ respectively denote a depthwise separable dilated convolution and a standard convolution, both with $k\times k$ kernel and $C$ output channels.
The computation process for $K_s$ and $V_s$ is similar, while their corresponding DD operators and standard convolutions update parameters independently. Subsequently, we obtain the channel attention map $A_s$ by flattened $query$ $\overline{Q}_s\in\mathds{R}^{C\times N}$ and $key$ $\overline{K}_s\in\mathds{R}^{C\times N}$, where $N=H\times W$, as follows:
\begin{equation}
\label{SDCA 2}
A_s=Sof(\overline{Q}_s \odot \overline{K}_s^T)\in\mathds{R}^{C\times C},
\end{equation}
where $Sof (\cdot)$ is the SoftMax function, and $\odot$ is matrix multiplication. 
Then, the activated attention map $A_s\in \mathds{R}^{C\times C}$ is used to recalibrate the parameters of flattened $value$ $\overline{V}_s$.
Finally, to enhance the local information in the spatial domain, we use depthwise separable dilated convolutions to modify the input feature $P$, and then concatenate the mappings to generate the spatial domain information $P_s \in\mathds{R}^{C\times H\times W}$, as shown in:
\begin{equation}
\label{SDCA 3}
P_s = \text{Conv}_1^C(\text{Cat}(P_s^s,P_s^r))\in\mathds{R}^{C\times H\times W},
\end{equation}
\begin{equation}
\label{SDCA 4}
P_s^s=\text{Reshape}(\text{Cat}(A_s\odot\overline{V_s})\in\mathds{R}^{C\times H\times W}, 
\end{equation}
\begin{equation}
\label{SDCA 5}
P_s^r=Cat(\text{DD}_3^\frac{C}{2}(P),\text{DD}_5^\frac{C}{2}(P))\in\mathds{R}^{C\times H\times W},
\end{equation}
where Reshape($\cdot$) refers to the inverse flattening operation, which adjusts the last dimension of the features into its original two-dimensional form suitable for convolution operations. $P_s^r$ is the spatial residual connection feature for providing and supplementing the original spatial information.

Compared to the input feature $P$, the output feature $P_s$ of SDCA block captures long-range channel dependencies and contains more informative content~\cite{SENet}, while also preserving the local details perceived in the spatial domain. This enhancement allows $P_s$ to provide a richer representation that effectively combines both broader contextual information and fine-grained local features.
\subsubsection{FDBA}
Specifically, as shown in~\cref{fig:overall architecture 2}, given the input feature $P\in\mathds{R}^{C\times H\times W}$, the Fourier transform is first applied to obtain its magnitude spectrum, which is then used as the $query$ $Q_f$, $key$ $K_f$, and $value$ $V_f$, as follows:
\begin{equation}
\label{FDBA 1}
Q_f=K_f=V_f=\mathcal{M}(\mathcal{F}(P))\in\mathds{R}^{C\times H\times W},
\end{equation}
Here, $\mathcal{F}(\cdot)$ and $\mathcal{M}(\cdot)$ represent the Fast Fourier Transform (FFT) and the magnitude computation formula, as shown in~\cref{fourier} and~\cref{magnitude}.
Unlike SDCA, we use the flattened $query$ $\overline{Q}_f\in\mathds{R}^{C\times N}$, $key$ $\overline{K}_f\in\mathds{R}^{C\times N}$, and $value$ $\overline{V}_f\in\mathds{R}^{C\times N}$, where $N=H\times W$, to obtain the spatial attention map $A_f$, as follows:
\begin{equation}
\label{FDBA 2}
A_f=Sof(\overline{Q}_f^T \odot \overline{K}_f)\in\mathds{R}^{N\times N},
\end{equation}
Subsequently, we use the spatial attention map to optimize the response values of different frequency bands in the magnitude map $\overline{V}_f$, adaptively enhancing the high-frequency features responsible for textures, as follows:
\begin{equation}
\label{FDBA 3}
\hat{P}_f^m=A_f \odot \overline{V}_f^T \in\mathds{R}^{N\times C}.
\end{equation}
Furthermore, to facilitate the inverse Fourier transform, we reconstruct it to obtain $P_f^m\in\mathds{R}^{C\times H\times W}$.

On the other hand, as shown in~\cref{fig:frequency analysis}, the phase spectrum is not sensitive to texture and contour information, but it contains shape and structural information and is numerically sensitive~\cite{DCT}. Therefore, we choose to retain it without additional feature extraction modules or other corrective operations, in order to accelerate the convergence of the model, as follows:
\begin{equation}
\label{FDBA 4}
P_f^p=\mathcal{P}(\mathcal{F}(P))\in\mathds{R}^{C\times H\times W},
\end{equation}
where $\mathcal{P}(\cdot)$ represents the phase computation operator, as shown in~\cref{phase}.
Then, we perform the inverse Fourier transform using the corrected magnitude spectrum and the original phase spectrum to obtain the corrected frequency domain feature $P_f^f$, as follows:
\begin{equation}
\label{FDBA 5}
P_f^f=\mathcal{F}^{-1}(P_f^m\cdot \text{exp}(jP_f^p))\in\mathds{R}^{C\times H\times W},
\end{equation}
where $\mathcal{F}^{-1}(\cdot)$ denotes the Invert Fast Fourier Transform (IFFT), as shown in~\cref{invert fourier}. In addition, we propose a frequency residual connection to enhance the frequency information, and combine the features to obtain the final frequency domain feature $P_f$, as follows:
\begin{equation}
\label{FDBA 6}
P_f=\text{Conv}_1^C(\text{Cat}(P_f^f,P_f^r))\in\mathds{R}^{C\times H\times W},
\end{equation}
\begin{equation}
\label{FDBA 7}
P_f^r=\mathcal{F}^{-1}(\sigma(\mathcal{F}(P)))\in\mathds{R}^{C\times H\times W},
\end{equation}
where $\sigma(\cdot)$ represents a sequence of operations, including a convolution, batch normalization, a ReLU function, another convolution, and a sigmoid function, utilized to obtain the frequency residual connection feature $P_f^r$ for providing and supplementing the original frequency information.

Compared to the input feature $P$, the output feature $P_f$ of FDBA block adaptively refines the magnitude spectrum through the self-attention mechanism, with the potential to enhance high-frequency information such as texture and contours in the foreground, while avoiding the introduction of background noise in the high-frequency domain.
\subsubsection{FSFN}
Frequency and spatial features typically focus on different aspects. The frequency domain focuses on the global energy distribution and variations of the signal, while spatial information deals with local pixel-level details and spatial structures.
Both provide valuable insights and clues for the overlapping object perception task, thus, the integration method of the two is crucial.
In our FSFN, these features are considered as two distinct states, which can undergo entanglement learning during the process to obtain more robust and powerful representations.

Specifically, FSFN consists of two parts. The first stage maps the input feature $P$ to both the frequency and spatial domains, enhancing the model's nonlinear representation capability using the GELU function, as well as employing a gating mechanism to retain global frequency features and local spatial information. The process is as follows:
\begin{equation}
\label{FSFN stage1 f}
P_c^f=GE(||\sigma( \mathcal{F}(P)\otimes \mathcal{F}(P))||)\otimes||\sigma( \mathcal{F}(P)\otimes \mathcal{F}(P))||,
\end{equation}
\begin{equation}
\label{FSFN stage1 s}
P_c^s=GE(\text{DD}_3^C(P))\otimes\text{DD}_3^C(P).
\end{equation}
Here, $GE(\cdot)$ denotes the GELU function, $\otimes$ represents the Hadamard product, and $||\cdot||$ denotes the modulus operation.

In the second stage, the frequency domain and spatial domain features from the first stage are first concatenated to obtain the joint feature $\hat{P}_c^c$ for feature interaction learning, and the process is as follows:
\begin{equation}
\label{FSFN 1}
\hat{P}_c^c=\text{Cat}(P_c^s,P_c^f).
\end{equation}
The feature $\hat{P}_c^c$ is fed into two branches for frequency domain learning and spatial domain learning, respectively. Information from both domains is interactively integrated from the perspectives of global energy perception and local detail perception, resulting in integrated features $\hat{P}_c^f$ and $\hat{P}_c^s$, which emphasize frequency domain and spatial domain information, respectively. Finally, these features are aggregated and the channels are reduced to form the combined feature $P_c$, as follows:
\begin{equation}
\label{FSFN 4}
P_c=\text{Conv}_1^C(\text{Cat}(\hat{P}_c^f,\hat{P}_c^s)),
\end{equation}
\begin{equation}
\label{FSFN 2}
\hat{P}_c^f=||\mathcal{F}^{-1}(\sigma(\mathcal{F}(\hat{P}_c^c))\otimes\mathcal{F}(\hat{P}_c^c)||, 
\end{equation}
\begin{equation}
\label{FSFN 3}
\quad\hat{P}_c^s=\text{DD}_3^C(\hat{P}_c^c).
\end{equation}
Compared to the input feature $P$, the output feature $P_c$ of FSFN performs entanglement learning on both spatial and frequency domain features, adaptively selecting and integrating them. It leverages the advantages of global frequency and local spatial information, resulting in a more comprehensive feature representation.

\section{Experiments}
\subsection{Implementation Details}
For fair comparisons, we train all models under the same conditions with the ImageNet~\cite{ImageNet} pretrained backbones, including ResNet-50~\cite{ResNet}, ResNet-101~\cite{ResNet}, ResNeXt-101~\cite{ResNeXt}, and Swin-L~\cite{Swin-Transformer}. CNN-based models are trained with SGD optimizer, using a learning rate of 0.01, momentum of 0.9, and weight decay of 0.1. Transformer-based models use the AdamW optimizer with a learning rate of 0.0001 and weight decay of 0.0001. All models are trained for 12 epochs and implemented using the MMDetection3.1.0 framework~\cite{MMDetection}, with an image size of $320 \times 320$.  All training is performed on a consistent computing platform equipped with an NVIDIA GeForce RTX 4090 GPU, an Intel Core i9-13900K CPU, 64 GB of memory, Windows 10 OS, and PyTorch 1.13.1. For the warm-up scheme in convolutional models, the learning rate increases linearly over the first 500 iterations with a warm-up ratio of 0.001. After this warm-up phase, the learning rate decreases stepwise, with adjustments made at the 8th and 11th epochs. For DETR-like models, following the approach of Deformable-DETR, the learning rate is reduced by a factor of 0.1 at the 11th epoch.

\subsection{Datasets and Evaluation Metrics}
\subsubsection{Datasets}
The PIXray~\cite{PIXray} dataset is capable of performing both object detection and instance segmentation tasks, referred to as PIXray-det and PIXray-seg, respectively. It includes 5046 X-ray images of prohibited items, divided into 4046 training images and 1000 testing images. It encompasses 15 categories of prohibited items, including Gun, Knife, Lighter, Battery, Pliers, Scissors, Wrench, Hammer, Screwdriver, Dart, Bat, Fireworks, Saw blade, Razor blade, and Pressure vessel. 

The OPIXray~\cite{OPIXray} is a fine-grained prohibited item dataset for sharp-edged tools, comprising 8885 X-ray images of prohibited items, allocated into 7019 training images and 1776 testing images. It includes five types of knives: Folding Knife (FO), Straight Knife (ST), Scissor (SC), Utility Knife (UK), Multi-tool Knife (MU).

The RSNA Lung Opacities (RLP) dataset is a subset of the pneumonia category data with fine-grained location labels that we have filtered from the RSNA Pneumonia Detection Challenge dataset~\cite{RSNA_Pneumonia_Detection_Challenge2018}. The original dataset had non-standardized labels. The RLP dataset is suitable for object detection training tasks. The dataset contains 6,011 X-ray images of pulmonary pneumonia, with 4,009 images used for training and 1,202 images used for testing. 
\subsubsection{Evaluation Metrics}
For the PIXray-det dataset and the RLP dataset, we apply the COCO~\cite{COCO} evaluation metrics. The main challenge metric is the box average precision (AP$^{box}$), which is calculated across 10 Intersection over Union (IoU) thresholds from 0.5 to 0.95, with a step of 0.05. Specifically, AP$^{box}_{50}$ indicates the mean average precision at an IoU threshold of 0.5, while AP$^{box}_{75}$ refers to the mean average precision at an IoU threshold of 0.75. In addition, AP$^{box}_S$, AP$^{box}_M$, and AP$^{box}_L$ correspond to the mean average precision for small objects ($\text{area} < 32^2$), medium objects ($32^2 < \text{area} < 96^2$), and large objects ($96^2 < \text{area}$), respectively.

For the OPIXray dataset, we use the VOC~\cite{VOC} evaluation metric. The average precision (AP) for each category is derived from the area under the Precision-Recall curve at an IoU threshold of 0.5. The mean average precision (mAP) is calculated by averaging the AP values across all categories. This mAP acts as a comprehensive evaluation metric, reflecting both the accuracy and recall of the detector, and offering a holistic view of its overall performance, including its strengths and weaknesses.

For the PIXray-seg dataset, the primary challenge metric is mask average precision (AP$^{mask}$), which is utilized to assess the performance of instance segmentation models comprehensively. It calculates the average precision (AP) based on different Intersection over Union (IoU) thresholds by computing the IoU between the predicted masks and the ground-truth masks. The distinction between it and AP$^{box}$ is merely in the objects that the IoU threshold is applied to. While AP$^{box}$ calculates the average precision using the IoU between bounding boxes, AP$^{mask}$ uses the IoU between predicted and ground truth masks, calculated pixel-wise.

\begin{table}
    \caption{Generalization analysis for FOAM on PIXray-det~\cite{PIXray} dataset for prohibited item detection.}
    \centering
    \begin{tabularx}{1\linewidth}{l|AAAAAA}
    \toprule
         Method&  AP$^{box}$ & AP$^{box}_{50}$ & AP$^{box}_{75}$ & AP$^{box}_S$ & AP$^{box}_M$ & AP$^{box}_L$ \\
         \midrule
         Deformable-DETR~\cite{Deformable-DETR}  & 36.6 &66.4 &37.1 &5.6 &24.2 &45.8 \\
         $\mathcal{F}$-Deformable-DETR  &\textbf{42.3} &70.3 &45.9 &5.5 &27.8 &50.8\\
         \midrule
         DINO~\cite{DINO}&64.3 &86.5 &71.0 &19.3 &48.9 &73.9\\
         $\mathcal{F}$-DINO&\textbf{67.2} &88.3 &74.8 &21.8 &52.6 &76.5\\
         \midrule
         RT-DETR~\cite{RT-DETR} &61.4 &84.0 &68.5 &20.0 &48.0 &70.1\\
         $\mathcal{F}$-RT-DETR &\textbf{61.7}&84.3 &68.6 &22.5 &48.1 &70.9\\
         \midrule
         Mask2Former$\dagger$~\cite{Mask2Former}&41.8 &61.8 &45.3 &8.2 &25.6 &51.7\\
         $\mathcal{F}$-Mask2Former$\dagger$&\textbf{43.0}&63.6 &46.6 &8.0 &25.1 &53.0\\
                  \midrule
         CondInst$\dagger$~\cite{CondInst}&53.4 &80.6 &60.6&12.7 &39.3 &62.4\\
         $\mathcal{F}$-CondInst$\dagger$&\textbf{54.4} &80.8 &61.7&12.5 &38.1 &63.5\\
                \midrule
         Cascade-Mask-R-CNN$\dagger$~\cite{Cascade_Mask_R-CNN}&70.2 &88.9 &78.9 &21.2 &58.5 &78.1\\
         $\mathcal{F}$-Cascade-Mask-R-CNN$\dagger$&\textbf{70.7} &88.8 &79.9 &20.1 &58.8 &78.6\\
                  \midrule
         Mask-R-CNN (X-101)$\dagger$~\cite{Mask}& 65.2 &88.6 &76.6 &11.5 &54.4 &73.2\\
         $\mathcal{F}$-Mask-R-CNN(X-101)$\dagger$&\textbf{65.6} &88.6 &77.4 &11.5 &54.5 &73.7 \\
          \midrule
         DINO (Swin-L)~\cite{DINO}&73.3 &90.2 &80.7 &39.4 &58.7 &80.9\\
        $\mathcal{F}$-DINO (Swin-L) &\textbf{73.7} &90.8 &80.3 &39.5 &60.8 &81.5\\
         \bottomrule
    \end{tabularx}
        \begin{flushleft}
$\dagger$ indicates models trained by both ``bounding box'' and ``segmentation'' labels, following~\cite{Mask,Cascade_Mask_R-CNN}. 

The default backbone of models is ResNet-50~\cite{ResNet}. X-101 and Swin-L stand for ResNeXt-101~\cite{ResNeXt} and Swin-Transformer-Large~\cite{Swin-Transformer} backbone, respectively.
\end{flushleft}
    \label{experiment Generalization analysis for FOAM on PIXray dataset for prohibited item detection.}
\end{table}

\begin{table*}[!t]
    \caption{Comparison with state-of-the-art object detectors on PIXray-det~\cite{PIXray} dataset.  }
    \centering 
    \begin{tabularx}{1\linewidth}{l|ccccc|AAAAAA}
    \toprule
    Method & Backbone & FPS & PARAMs (M) & FLOPs (G) & \# queries & AP$^{box}$ & AP$^{box}_{50}$ & AP$^{box}_{75}$ & AP$^{box}_S$ & AP$^{box}_M$ & AP$^{box}_L$\\
    \midrule
        \multicolumn{12}{c}{General Object Detectors}\\
    \midrule
    Faster R-CNN~\cite{Faster}& ResNeXt-101&70& 59.83 & 28.35&*& 53.6 & 82.3 & 60.8 & 3.9 & 37.7 & 62.7 \\
    Cascade R-CNN~\cite{Cascade}&ResNet-50&39&69.20&60.99&*&61.0 &83.9 &69.0 &10.4 &46.8 &69.7\\
    ATSS~\cite{ATSS}&ResNet-101&66&51.14&27.82&*&52.8&80.8&60.2&7.0&37.4&63.6\\
    GFLv1~\cite{GFLv1}&ResNeXt-101&66&50.70&28.51&*&57.5&82.8&66.0&9.1&42.0&67.4\\
    DINO~\cite{DINO}&ResNet-50 & 54 & 58.38 & 26.89 &  30 & 64.3 & 86.5 & 71.0 & 19.3 & 48.9 & 73.9\\
    RT-DETR~\cite{RT-DETR}& ResNet-50 & 64 & 42.81 & 17.07& 60 & 62.3 & 85.3 & 69.9 & 25.6 & 48.0 &70.9\\
    DINO~\cite{DINO}&Swin-L & 40 &229.0& 156.0&  30 &73.3 &90.2 &\textbf{80.7} &39.4 &58.7 &80.9\\
    \midrule
        \multicolumn{12}{c}{Prohibited Item Detectors}\\
    \midrule
    AO-DETR~\cite{AO-DETR}& ResNet-50 & 54 & 58.38 & 26.89& 30 & 65.6 & 86.1 & 72.0 & 23.9 & 50.7 & 74.8\\
    $\mathcal{M}$-DINO~\cite{MMCL}&ResNet-50& 54 & 58.38 & 26.89& 30 & 66.7 & 87.5 & 74.4 & 23.5 & 50.7 &75.5\\
    $\mathcal{M}$-RT-DETR~\cite{MMCL}& ResNet-50 & 64 & 42.81 & 17.07& 60 & 63.6 & 85.9 & 71.4 & 24.0 & 49.9 &72.6\\
    $\mathcal{C}$-DINO~\cite{CSPCL}& ResNet-50&54 & 58.38 & 26.89& 30 &66.4 &86.8 &73.6 &25.7 &50.9 &75.6\\
    $\mathcal{C}$-RT-DETR~\cite{CSPCL}&ResNet-50& 64 & 42.81 & 17.07&30&61.8 &84.3 &68.7 &25.2 &47.7 &70.6\\
    \midrule
    $\mathcal{F}$-DINO (Ours)&ResNet-50& 38&59.75&30.79&30 &67.2 &88.3 &74.8 &21.8 &52.6 &76.5\\
    $\mathcal{F}$-DINO (Ours) &Swin-L &29&230.37 &171.93& 30 &\textbf{73.7} &\textbf{90.8} &80.3 &\textbf{39.5} &\textbf{60.8} &\textbf{81.5}\\
    \bottomrule
    \end{tabularx}
    \label{experiment Comparison with State-of-the-Art Methods results on pixray}
\end{table*}   
\begin{table*}[!t]
    \caption{Comparison with state-of-the-art object detectors on OPIXray~\cite{OPIXray} dataset. FO, ST, SC, UT, and MU represent Folding Knife, Straight Knife, Utility Knife, and Multi-Tool Knife, respectively.  }
    \centering 
    \begin{tabularx}{1\linewidth}{l|ccccc|AAAAAA}
    \toprule
    Method & Backbone & FPS & PARAMs (M) & FLOPs (G) & \# queries& mAP&FO& ST& SC& UT& MU\\
    \midrule
        \multicolumn{12}{c}{General Object Detectors}\\
    \midrule
    Faster R-CNN~\cite{Faster}& ResNeXt-101&70& 59.83 & 28.35&*& 73.4&80.6 & 45.4 & 89.1  & 69.1 & 83.1 \\
    Cascade R-CNN~\cite{Cascade}&ResNet-50&39&69.20&60.99&*&76.9 &83.8& 58.8& 90.0 &73.2 &78.8\\
    ATSS~\cite{ATSS}&ResNet-101&66&51.14&27.82&*&67.5&72.8&38.0&88.6&58.0&80.2\\
    GFLv1~\cite{GFLv1}&ResNeXt-101&66&50.70&28.51&*&75.6&80.0&53.6&89.3&71.7&83.4\\
    DINO~\cite{DINO}&ResNet-50 & 54 & 58.38 & 30.79 &  30 & 78.2 & 83.2 & 58.8 & 89.4 & 72.7 & 86.7\\
    RT-DETR~\cite{RT-DETR}& ResNet-50 & 64 & 42.81 & 17.07& 320 & 61.8 & 61.1 & 26.0 & 88.6 & 56.4 &76.8\\
    DINO~\cite{DINO}&Swin-L & 40 &229.0& 156.0&  30 & 80.0& 84.2& 61.1& 89.0 &\textbf{78.9}& 86.6\\
    \midrule
        \multicolumn{12}{c}{Prohibited Item Detectors}\\
    \midrule
    AO-DETR~\cite{AO-DETR}& ResNet-50 & 54 & 58.38 & 26.89& 30 & 79.2 & 83.8 & 60.5 & 90.1 & 74.7 & 87.1\\
    $\mathcal{M}$-DINO~\cite{MMCL}&ResNet-50& 54 & 58.38 & 26.89& 30 & 78.6 & 83.9 & 57.2 & \textbf{90.4} & 74.2 &87.1\\
    $\mathcal{M}$-RT-DETR~\cite{MMCL}& ResNet-50 & 64 & 42.81 & 17.07& 320 & 62.5 & 65.9 & 22.3 & 86.4 & 57.1 &80.7\\
    $\mathcal{C}$-DINO~\cite{CSPCL}& ResNet-50&54 & 58.38 & 26.89& 30 &77.9 &82.8 &56.0 &89.9 &74.2 &86.7\\
    $\mathcal{C}$-RT-DETR~\cite{CSPCL}&ResNet-50& 64 & 42.81 & 17.07&30&70.1&76.0 &34.4 &88.6 &67.4 &84.3\\
    \midrule
    $\mathcal{F}$-DINO (Ours)&ResNet-50& 38&59.75&30.79&30 &79.8 &84.3 &62.3 &89.9 &74.9 &87.5\\
    $\mathcal{F}$-DINO (Ours) &Swin-L &29&230.37 &171.93& 30 &\textbf{81.7} &\textbf{86.4} &\textbf{65.2} &89.3 &78.6 &\textbf{89.0}\\
    \bottomrule
    \end{tabularx}
    \label{experiment Comparison with State-of-the-Art Methods results on opixray}
\end{table*} 
\begin{table*}
    \caption{Comparison with state-of-the-art instance segmentation models on PIXray-seg~\cite{PIXray} dataset.}
    \centering
    \begin{tabularx}{1\linewidth}{l|ccccc|AAAAAA}
    \toprule
         Instance Segmentation& Backbone & FPS & PARAMs (M) & FLOPs (G) & \# queries & AP$^{mask}$ & AP$^{mask}_{50}$ & AP$^{mask}_{75}$ & AP$^{mask}_S$ & AP$^{mask}_M$ & AP$^{mask}_L$ \\
         \midrule

         Mask2Former$\dagger$~\cite{Mask2Former}&ResNet-50&12&176.11& 26.79&100&37.7 &66.1 &37.2 &6.3 &18.8 &50.2\\
         SOLO~\cite{SOLO}&ResNet-50&38&36.15&51.23&*&33.5 &65.0 &31.0&0.8 &15.2 &44.3\\
         SOLOv2~\cite{SOLOv2}&ResNet-50&24&46.29&86.37&*&35.2 &68.5 &32.3 &0.1 &16.6 &47.2\\
         CondInst$\dagger$~\cite{CondInst}&ResNet-50&23&34.01&33.60&*&30.3 &62.5 &26.8 &1.9 &14.9 &37.8\\
         Cascade-Mask-R-CNN$\dagger$~\cite{Cascade_Mask_R-CNN}&ResNet-50&44&77.08&1271.85&*&55.1 &85.7 &60.2 &\textbf{9.5} &37.3 &62.8\\
         Mask-R-CNN$\dagger$~\cite{Mask}&ResNeXt-101&30&107.31&110.12&*&55.2 &85.8 &61.1 &3.9 &38.0 &63.5\\
         $\mathcal{F}$-Mask-R-CNN$\dagger$ (Ours)&ResNeXt-101&26&109.48&112.08&*&\textbf{55.8} &\textbf{86.8} &\textbf{61.5} &5.3 &\textbf{39.2} &\textbf{63.9}\\
         \bottomrule
    \end{tabularx}
    \begin{flushleft}
$\dagger$ indicates models trained by both ``bounding box'' and ``segmentation'' labels, following~\cite{Mask,Cascade}.
\end{flushleft}
    \label{experiment Comparison with State-of-the-Art Methods results for prohibited item segmentation}
\end{table*}
\begin{table*}[!t]
    \caption{Comparison with state-of-the-art object detectors on RLP dataset.  }
    \centering 
    \begin{tabularx}{1\linewidth}{l|ccccc|AAAAAA}
    \toprule
    Method & Backbone & FPS & PARAMs (M) & FLOPs (G) & \# queries & AP$^{box}$ & AP$^{box}_{50}$ & AP$^{box}_{75}$ & AP$^{box}_S$ & AP$^{box}_M$ & AP$^{box}_L$\\
    \midrule

    RT-DETR~\cite{RT-DETR}& ResNet-50 & 64 & 42.81 & 17.07& 60 &14.7 &38.9 &8.0 &2.6 &12.7 &21.3\\
    DINO~\cite{DINO}&ResNet-50 & 54 & 58.38 & 26.89 &  30 &15.6 &44.7 &6.6 &5.4 &13.1 &25.0\\
    DINO~\cite{DINO}&Swin-L & 40 &229.0& 156.0& 30 &16.3 &44.6 &7.8 &4.4 &12.7 &25.9\\
    ATSS~\cite{ATSS}&ResNet-101&66&51.14&27.82&*&16.1&48.9 &5.0 &0.3 &12.7 &25.0\\
    GFLv1~\cite{GFLv1}&ResNeXt-101&66&50.70&28.51&*&10.2 &35.4 &1.7 &0.0 &9.5 &14.8\\
    Faster R-CNN~\cite{Faster}& ResNeXt-101&70& 59.83 & 28.35&*& 18.6 &55.7 &7.0 &\textbf{4.1} &15.9 &26.3 \\
    Cascade-R-CNN~\cite{Cascade}&ResNet-50&39&69.20&60.99&*&19.7 &\textbf{56.8} &8.0 &2.5 &16.3 &28.3\\
    $\mathcal{F}$-Cascade-R-CNN (Ours) &ResNet-50&36&70.36&72.96&*&\textbf{20.2} &56.7 &\textbf{8.6} &2.1 &\textbf{16.4} &\textbf{29.2}\\
    \bottomrule
    \end{tabularx}
    \label{experiment Comparison with State-of-the-Art Methods results on RLP}
\end{table*}
\subsection{Generalization}
\subsubsection{Models and Backbones}
In this part, we first demonstrate the powerful architectural and model generalization capabilities of FOAM by applying it to various object detectors under two advanced architectures. We then select representative models with strong backbones, including the convolutional-based ResNeXt-101~\cite{ResNeXt} and the transformer-based Swin-L~\cite{Swin-Transformer}, to further validate the backbone generalization ability of FOAM. As shown in~\cref{experiment Generalization analysis for FOAM on PIXray dataset for prohibited item detection.}, FOAM achieves box AP gains of $4.3\%$, $2.9\%$, $0.3\%$, and $1.2\%$ for Deformable-DETR~\cite{Deformable-DETR}, DINO~\cite{DINO}, RT-DETR~\cite{RT-DETR}, and Mask2Former~\cite{Mask2Former} on the PIXray-det~\cite{PIXray} dataset, respectively, highlighting its effectiveness for emerging and advanced Deformable-DETR-based models. Similarly, FOAM improves the box AP by $0.5\%$, $0.4\%$, and $1.0\%$ for Cascade-Mask-R-CNN~\cite{Cascade_Mask_R-CNN}, Mask-R-CNN~\cite{Mask}, and CondInst~\cite{CondInst}, respectively, demonstrating its effectiveness for traditional models based on the fully convolutional architecture. Finally, FOAM boosts the box AP by $2.9\%$ and $0.4\%$ for DINO with ResNet-50 and Swin-L backbones, and by $0.4\%$ for Mask-R-CNN with ResNeXt-101, illustrating its strong generalization capability across both CNN- and Transformer-based backbones.
\subsection{Comparison with SOTA Models}
To further validate the improvement in overlapping object perception brought by the supplementary frequency domain information from FOAM, we leverage FOAM across four datasets to challenge state-of-the-art algorithms on three overlapping object perception tasks, including Prohibited Item Detection, Prohibited Item Segmentation, and Pneumonia Detection.
\subsubsection{Experiments over Prohibited Item Detection}
We challenge state-of-the-art models in the prohibited item detection domain on the PIXray-det~\cite{PIXray} and OPIXray~\cite{OPIXray} datasets.

The quantitative results for the PIXray-det~\cite{PIXray} dataset are presented in~\cref{experiment Comparison with State-of-the-Art Methods results on pixray}. Notably, with the same ResNet-50 backbone, $\mathcal{F}$-DINO, which is the combination of FOAM and DINO, outperforms other DINO-based improved prohibited item detectors, including AO-DETR~\cite{AO-DETR}, $\mathcal{M}$-DINO~\cite{MMCL}, and $\mathcal{C}$-DINO~\cite{CSPCL}, in terms of accuracy. This demonstrates that the FOAM architecture enhances the model's ability to perceive foreground features from overlapping scenes more effectively than other methods in the prohibited item detection domain, such as CSA~\cite{AO-DETR}, MMCL~\cite{MMCL}, and CSPCL~\cite{CSPCL}.
The version of the small-scale model, $\mathcal{F}$-DINO (ResNet-50) achieves the best performance with a box AP of $67.2\%$, among models with comparable parameters and FLOPs.
For the version of the large-scale model, $\mathcal{F}$-DINO (Swin-L) achieves $73.7\%$ box AP, surpassing both general object detectors and specialized prohibited item detectors.

In addition, we evaluate the performance of FOAM on the fine-grained sharp-edged tools dataset, OPIXray~\cite{OPIXray}. As shown in~\cref{experiment Comparison with State-of-the-Art Methods results on opixray}, $\mathcal{F}$-DINO achieves higher mAP ($79.8\%$) than AO-DETR, $\mathcal{M}$-DINO, and $\mathcal{C}$-DINO, demonstrating that the additional frequency domain cues provided by FOAM enable the model to more effectively understand and distinguish subtle foreground differences in overlapping scenes compared to other anti-overlapping strategies.
For the version of the large-scale model, $\mathcal{F}$-DINO (Swin-L) achieves an mAP of $81.7\%$, exceeding other state-of-the-art models in both the fields of general object detection and prohibited item detection.
\subsubsection{Experiments over Prohibited Item Segmentation}
To explore the effectiveness of FOAM in instance segmentation tasks under overlapping scenes.
We challenge state-of-the-art models in prohibited item segmentation on the PIXray-seg~\cite{PIXray} dataset. As shown in~\cref{experiment Comparison with State-of-the-Art Methods results for prohibited item segmentation}, Mask-R-CNN (ResNeXt-101)~\cite{Mask} achieves a mask AP of $55.2\%$, outperforming other instance segmentation models, such as Cascade-Mask-R-CNN~\cite{Cascade_Mask_R-CNN}. Building upon this, $\mathcal{F}$-Mask-R-CNN further improves the mask AP to $55.8\%$, surpassing other instance segmentation models, including DETR-based Mask2Former~\cite{Mask2Former} and fully convolutional models such as SOLO~\cite{SOLO}, SOLOv2~\cite{SOLOv2}, and CondInst~\cite{CondInst}.
\subsubsection{Experiments over Pneumonia Detection}
To further explore the application of FOAM, we are the first to apply anti-overlapping detection techniques in the medical diagnostic field. Specifically, we applied FOAM to the pneumonia detection dataset, RLP, based on X-ray images. As shown in~\cref{experiment Comparison with State-of-the-Art Methods results on RLP}, traditional multi-stage or two-stage object detectors, such as Cascade-R-CNN and Faster-R-CNN, achieve better accuracy compared to single-stage detectors like ATSS and GFLv1, and outperform Deformable-DETR-based models like RT-DETR and DINO, which perform better on the general object detection dataset COCO~\cite{COCO}. Therefore, we utilize the Cascade-R-CNN as the baseline model, and observe that under the influence of FOAM, $\mathcal{F}$-Cascade-R-CNN improves the box AP from $19.7\%$ to $20.2\%$, demonstrating the strong generalization ability of FOAM.
\subsection{Ablation Study}
In this part, to optimize the proposed method FOAM, we conducted extensive ablation experiments using DINO as the baseline on the PIXray-det dataset.
\subsubsection{ Ablation study for HDC and FSTB}
\begin{table}
    \caption{Ablation study of HDC and FSTB results. PARAMs, FLOPs, and FPS represent the total number of parameters, floating point operations, and the number of inferences the model can perform per second, respectively.}
    \centering
    \begin{tabularx}{1\linewidth}{A|AA|c|c|A|A}
    \toprule
         N&FSTB &HDC&PARAMs(M)&FLOPs(G)&FPS&AP$^{box}$ \\
         \midrule
         0&\ding{55}&\ding{55}&58.380& 26.820&54&64.3 \\
         \midrule
         \multirow{2}*{1}
          &\ding{51}&\ding{55}&59.746& 30.791&38&66.0\\
         ~&\ding{51}&\ding{51}&59.746& 30.791&38&67.2\\
                  \midrule
         \multirow{2}*{2}
          &\ding{51}&\ding{55}&61.111& 34.762&30&66.4\\
         ~&\ding{51}&\ding{51}&61.111& 34.762&30&67.3\\
                  \midrule
         \multirow{2}*{3}
          &\ding{51}&\ding{55}&62.477& 38.732&23&66.9\\
         ~&\ding{51}&\ding{51}&62.477& 38.732&23&67.7\\
         \bottomrule
    \end{tabularx}
    \label{experiment Ablation study of HDC and FSTB results.}
\end{table}
\cref{experiment Ablation study of HDC and FSTB results.} presents a complex ablation study that thoroughly evaluates the effects of the Frequency Spatial Transformer Block (FSTB), the number of iterations $N$ of cascaded FSTBs for corruption and base branches, and the Hierarchical De-Corrupting (HDC) mechanism on the model's performance. When the number of iterations $N=1$, the FSTB is able to increase the model's box AP from $64.3\%$ to $66.0\%$. Furthermore, when both the HDC and FSTB are utilized together, the box AP reaches $67.2\%$, indicating good compatibility between the two methods and demonstrating that the HDC mechanism effectively guides the FSTB to achieve a stronger anti-overlapping detection capability. Additionally, the PARAMs and FLOPs only increased by $1.366$ M and $3.971$ G, while FPS decreased by $16$ frames. This suggests that the method is not demanding in terms of computational resources. Similarly, when we increase the number of iterations $N$ to $2$ and $3$, FSTB and HDC continue to show good compatibility, and the model's box AP improved further, albeit with diminishing returns. To balance computational complexity and performance, we set $N$ to $1$ for subsequent experiments.

\begin{table}
    \caption{
    Comparison of Gaussian Blurring (GB), Downsampling and Upsampling (DU), and Gaussian Noise (GN) corruption strategies for the HDC mechanism. ks means the kernel size of Gaussian Noise}
    \centering
    \begin{tabularx}{1\linewidth}{l|c|AAAAAA}
    \toprule
         Method& Parameter &AP$^{box}$ & AP$^{box}_{50}$ & AP$^{box}_{75}$ & AP$^{box}_S$ & AP$^{box}_M$ & AP$^{box}_L$ \\
         \midrule
         --&--&64.3 & 86.5 & 71.0 & 19.3 & 48.9 & 73.9\\
         \midrule
         \multirow{4}*{(a) DU}&$\times 2$&65.6 &87.1 &72.9 &18.7 &50.9 &75.4\\
         ~&$\times 3$&65.3 &86.9 &72.5 &20.6 &50.2 &74.7\\
         ~&$\times 4$&\textbf{66.5} &88.0 &73.3 &25.0 &51.9 &75.9\\
         ~&$\times 5$&64.8 &86.6 &71.8 &19.5 &50.2 &74.3\\
         \midrule
         \midrule
          \multirow{4}*{(b) GN}&0.1&65.7 &87.3 &73.5 &21.1 &51.4 &75.1\\
         ~&0.2&\textbf{66.4} &87.9 &73.8 &23.0 &52.3 &75.6\\
         ~&0.5&66.1 &87.7 &72.7 &21.7 &50.8 &75.9\\
         ~&1&65.4 &87.5 &71.9 &21.5 &51.5 &74.8\\
                  \midrule
         \midrule
          \multirow{4}*{(c) GB (ks=3)}&0.1&66.6 &87.9 &73.5 &21.2 &51.9 &75.7\\
         ~&1&66.9 &88.5 &73.8 &23.6 &51.1 &76.4\\
         ~&5 &\textbf{67.2} &88.3 &74.8 &21.8 &52.6 &76.5\\
         ~&10 &66.5 &88.2 &73.9 &23.0 &50.8 &76.2\\
                  \midrule
         \midrule
          \multirow{4}*{(d) GB (ks=5)}&0.1&63.9 &86.0 &70.6 &21.8 &48.4 &73.0\\
         ~&1&66.4 &87.5 &73.6 &22.0 &51.9 &75.9\\
         ~&5&\textbf{66.5} &87.7 &74.2 &21.3 &51.6 &76.1\\
         ~&10&65.5 &86.9 &72.0 &22.1 &50.4 &75.0\\
         \bottomrule
    \end{tabularx}
    \begin{flushleft}
The studied hyperparameters that control the degradation level of each corruption strategy are (a) the Downsampling scale factor, (b) the Gaussian Noise sigma, and (c) the Gaussian Blurring sigma.
\end{flushleft}
    \label{experiment Comparison of corruption strategies}
\end{table}
\subsubsection{Ablation study for corruption strategies}
As shown in~\cref{experiment Comparison of corruption strategies}, to further investigate the effects of different corruption strategies, such as Gaussian Blurring (GB), Downsampling and Upsampling (DU), and Gaussian Noise (GN), on the guidance provided by the HDC mechanism for enhancing the texture and contour feature perception capabilities of FSTB, we compare the performance of the $\mathcal{F}$-DINO model on the PIXray-det dataset under various hyperparameters for each strategy. For the DU strategy, we adjust the Downsampling scale factor and find that a factor of “×4” achieves the highest $66.5\%$ box AP compared to other coefficients. Under the Gaussian Noise (GN) strategy, when the Gaussian Noise sigma is set to $0.2$, the model gains an increase of $2.1\%$ box AP. In the case of the GB strategy, the models' overall performances are superior when the Gaussian kernel size (ks) is set to $3$ as opposed to $5$. Furthermore, when $ks=3$ and the Gaussian Blurring sigma is set to $5$, the model achieved its highest box AP of $67.2\%$. Overall, the GB strategy provides the most positive impact for the anti-overlapping feature-awareness capability of the FSTB module.
\subsubsection{Ablation study for consistent losses}\label{sec:Ablation study for consistent losses}
\begin{figure}
    \centering
    \includegraphics[width=1\linewidth]{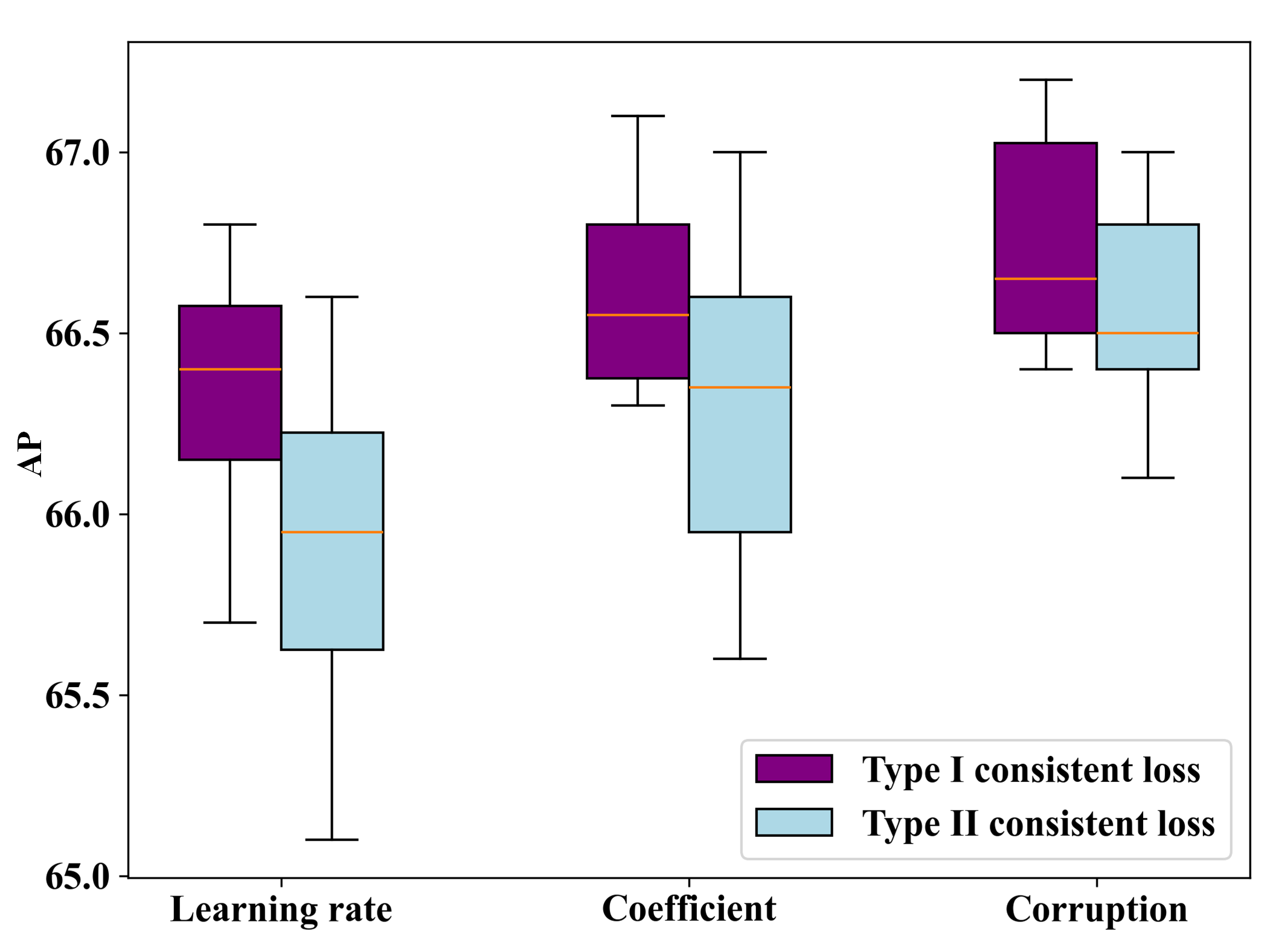}
    \caption{ Comparison of Type I consistent Loss and Type II consistent Loss. The standard boxplot illustrates the relationship between box AP and variations in the learning rates, coefficients of consistent loss, and corruption strategies.}
    \label{fig:boxplot}
\end{figure}

As shown in~\cref{fig:boxplot}, we conduct three sets of experiments, each using different learning rates, loss weighting coefficients, and corruption strategies to compare the impact of Type I consistent loss, as defined in~\cref{equation KL loss}, versus Type II consistent loss, as defined in~\cref{equation MSE loss}, on the box AP of $\mathcal{F}$-DINO on the PIXray dataset, whose results are presented as the boxplot.
All experiments are conducted with the same batch size, learning strategy, data augmentation strategy, and optimizer to exclude potential interference from other factors. The experimental results are progressively improved, as each subsequent experimental group uses the best parameter settings from the previous group. For example, the corruption control group uses the optimal learning rate and coefficient settings from the learning rate and coefficient control groups, respectively.
It can be observed that although both loss functions positively impact the model's accuracy, which exceeds the baseline box AP of $64.3\%$, the mean and maximum of the outputs for Type I consistent loss based on KL divergence are higher, while its variance is overall lower compared to Type II consistent loss based on MSE. This discrepancy is attributed to the fact that Type I loss emphasizes the relative difference between $F_C^{n+1}$ and $F_O^n$, whereas Type II loss focuses on the absolute difference between the two. Therefore, Type I loss is more robust and effective for fine-grained feature-level alignment tasks.
\begin{table}
        \caption{Ablation study of the target layer set $\mathcal{L}$ under the Gaussian Blurring strategy (kernel size is 5, sigma is $5$).}
    \centering
    \begin{tabularx}{1\linewidth}{AAAA|AAA}
    \toprule
        $l=1$& $l=2$ & $l=3$ & $l=4$ & AP$^{box}$ & AP$^{box}_{50}$ & AP$^{box}_{75}$\\
         \midrule
         \ding{55}&\ding{55}&\ding{55}&\ding{55}&64.3 & 86.5 & 71.0\\
         \ding{55}&\ding{55}&\ding{55}&\ding{51}&67.1 &88.1 &74.8 \\
         \ding{55}&\ding{55}&\ding{51}&\ding{51}&\textbf{67.2} &88.3 &74.8 \\
         \ding{55}&\ding{51}&\ding{51}&\ding{51}&66.9 &88.0 &74.3 \\
         \ding{51}&\ding{51}&\ding{51}&\ding{51}&65.2 &87.8 &72.7 \\
         \bottomrule
    \end{tabularx}
    \label{experiment Ablation study of the target layer set.}
\end{table}
\begin{table}
    \caption{Ablation study of SDCA and FDBA.}
    \centering
    \begin{tabularx}{1\linewidth}{c|c|c|c|AAA}
    \toprule
        \multirow{2}*{ID}&SDCA& \multicolumn{2}{c|}{FDBA} &  \multirow{2}*{AP$^{box}$} & \multirow{2}*{AP$^{box}_{50}$} & \multirow{2}*{AP$^{box}_{75}$}\\
        \cmidrule{2-4}
        ~&Shape of $A_s$ &Shape of $A_f$ &Target&~&~&~\\
         \midrule
         (a)&C$\times$C&N$\times$N&Magnitude&\textbf{67.2} &88.3 &74.8 \\
         (b)&N$\times$N&N$\times$N&Magnitude&67.0 &87.8 &74.5\\
         \midrule
         (c)&C$\times$C&C$\times$C&Magnitude&66.0 &87.2 &74.6 \\
         (d)&C$\times$C&C$\times$C&Phase&65.2 &86.5 &73.7 \\ 
         (e)&C$\times$C&N$\times$N&Phase&65.8 &87.2 &74.2 \\
         \bottomrule
    \end{tabularx}
\begin{flushleft}
$A_s$ means channel attention map in~\cref{SDCA 2}, and $A_f$ means spatial attention map in~\cref{FDBA 2}.
\end{flushleft}
    \label{experiment Ablation study of SDCA and FDBA.}
\end{table}
\subsubsection{Ablation study for target layer set}\label{sec:Ablation study for target layer set}
Since $F_C^{n+1}=\{F_{C,l}^{n+1}\}_{l=0}^L$ and $F_O^n=\{F_{O,l}^n\}_{l=0}^L$ are multi-scale features with $L=4$ layers, we conduct experiments on $\mathcal{F}$-DINO to investigate which feature layers in~\cref{equation KL loss} need to be aligned using the consistent loss mechanism in the HDC mechanism to better guide the FSTB. The results show that the model achieves higher accuracy when consistency guidance is applied for higher-level features.
This can be attributed to the fact that higher-level features primarily capture global information, which is more adept at encoding semantic information. In contrast, lower-level features tend to focus on local information, often accompanied by redundant data and noise. The presence of ineffective information from these lower-level features disrupts the FSTB’s ability to comprehend the reverse process of corruption during consistency training, thereby reducing its ability to extract features in the presence of overlap.
\subsubsection{Ablation study of SDCA and FDBA}\label{sec:Ablation study of SDCA and FDBA}
To enable the FSTB to optimize and maximize the integration of local spatial features and global frequency domain information, we conduct a series of ablation experiments, as shown in~\cref{experiment Ablation study of SDCA and FDBA.}.
By comparing~\cref{experiment Ablation study of SDCA and FDBA.}(a) and~\cref{experiment Ablation study of SDCA and FDBA.}(b), we observe that the Spatial Domain Channel Self-Attention (SDCA) mechanism is better suited for using a channel attention matrix of the form $A_s \in\mathds{R}^{C\times C}$, rather than the classical spatial domain feature reorganization with an attention map of the form $A_s \in\mathds{R}^{N\times N}$~\cite{Attention_is_all_you_need}, where $N=H\times W $ represents the flattened spatial scale.
By comparing~\cref{experiment Ablation study of SDCA and FDBA.}(c) and~\cref{experiment Ablation study of SDCA and FDBA.}(d), we find that the Frequency Domain Bands Self-Attention (FDBA) mechanism is better suited for adaptively correcting the magnitude spectrum, which is more effective at representing texture and contour information, rather than the phase spectrum, which excels at capturing shape and structural information.
Finally, by comparing~\cref{experiment Ablation study of SDCA and FDBA.}(d) with~\cref{experiment Ablation study of SDCA and FDBA.}(e), or~\cref{experiment Ablation study of SDCA and FDBA.}(a) with~\cref{experiment Ablation study of SDCA and FDBA.}(c), we observe that for FDBA, the classical spatial attention mechanism with $A_f \in\mathds{R}^{N\times N}$ is more suitable. This is because this approach is better at integrating effective information from both the high-frequency and low-frequency bands of the magnitude spectrum.

Overall, the configuration in~\cref{experiment Ablation study of SDCA and FDBA.}(a), as described in~\cref{sec:FSTB}, achieves the best accuracy. In this configuration, SDCA preserves the local features that spatial domain features excel at, while also capturing long-range channel dependencies and containing more informative content~\cite{SENet}. Meanwhile, FDBA integrates both high-frequency and low-frequency information from the magnitude spectrum, particularly regarding texture and contour. By leveraging the complementary characteristics of both feature domains, the system incorporates both global and local features, providing high-quality input features for the subsequent coupling learning in FSFN.
\subsection{Visualization and Analysis}
In this part, we first use the state-of-the-art model, DINO, on PIXray-det as the baseline and conduct a progressive analysis of the impact of FOAM across three levels: feature extraction by the backbone network, feature extraction in the decoder, and high-order statistical analysis of the final inference results. This includes visualizing feature maps to assess how FOAM influences the feature extraction process of the backbone, examining the decoder of the last layer via sampling and reference points, and utilizing scatter plots to analyze classification and localization results.
Finally, we visualize the prediction results of the SOTA models on four datasets for the three tasks, qualitatively comparing the impact of FOAM on the prediction outcomes of models.
\begin{figure}
    \centering
    \includegraphics[width=1\linewidth]{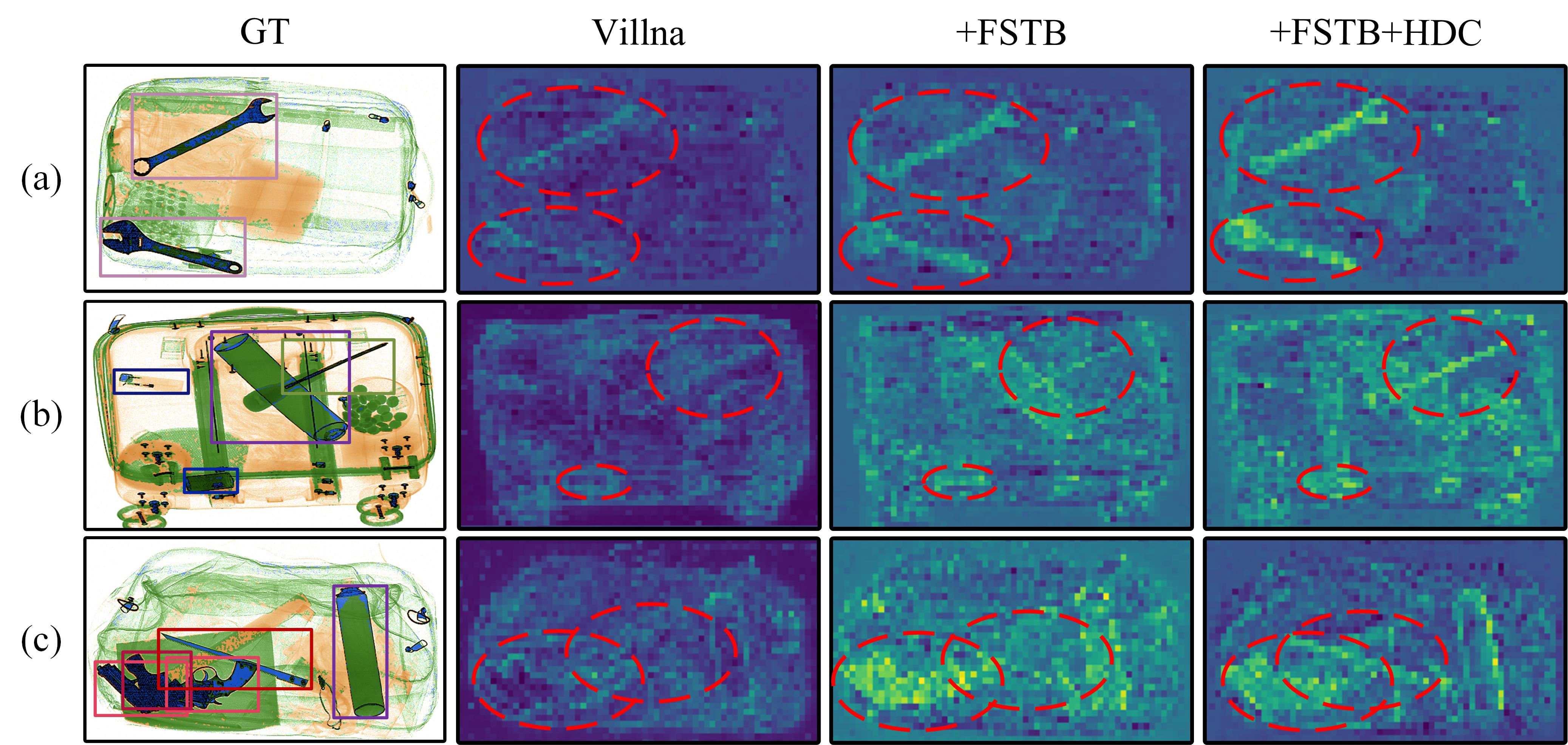}
    \caption{
    Visualization of feature maps of ``DINO'', ``DINO+FSTB'', and ``DINO+FSTB+HDC'' ($\mathcal{F}$-DINO).
    }
    \label{fig:attention map visualization3}
\end{figure}
\begin{figure}
    \centering
    \includegraphics[width=1\linewidth]{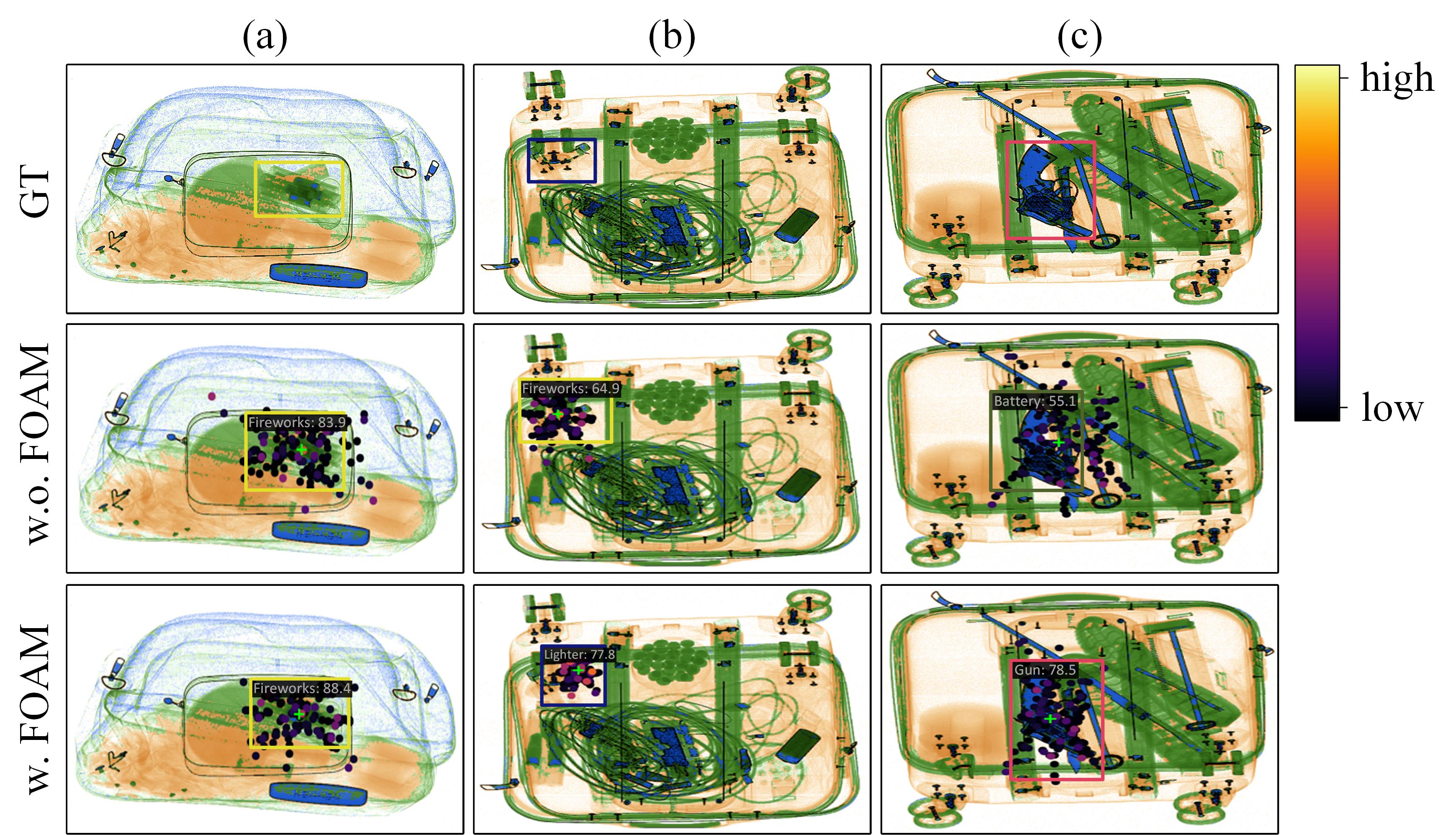}
    \caption{Visualization of deformable attention sampling points, reference points, and prediction results for corresponding content query in the last decoder layer. Row ``GT'', ``w.o. FOAM'', and ``w. FOAM'' refers to the ground truth, and results of DINO and $\mathcal{F}$-DINO. Column ``(a)-(c)'' represents images of PIXray-det~\cite{PIXray} dataset. Each sampling point is shown as a filled circle, with color indicating its attention weight, and the reference point is marked by a green cross.}
    \label{fig:sampling points}
\end{figure}
\begin{figure}
    \centering
    \includegraphics[width=1\linewidth]{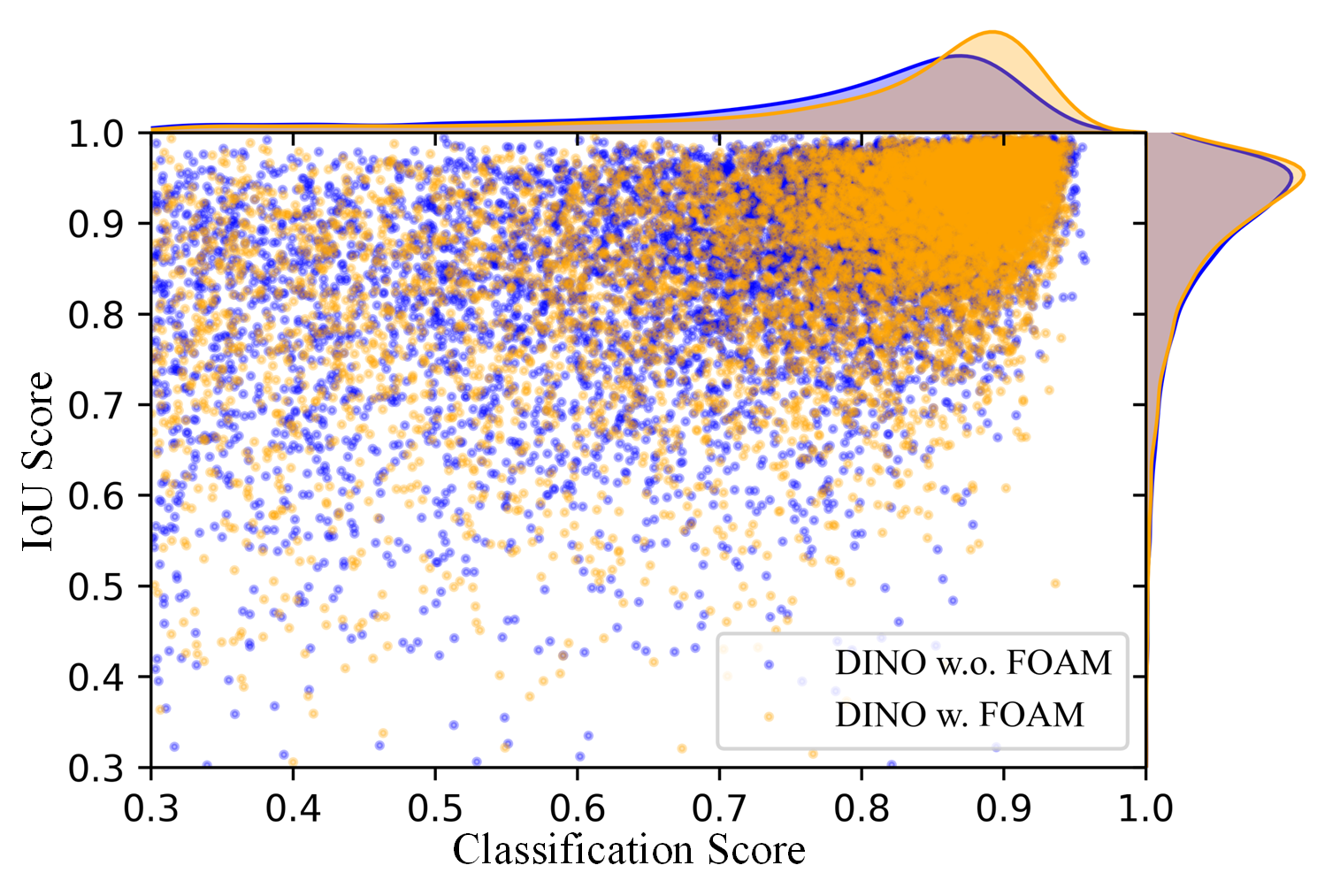}
    \caption{The scatter plot and Kernel Density Estimation (KDE) joint distribution plot of the prediction results from the final decoder layer. \textcolor{bluejoint}{Blue} and \textcolor{orangejoint}{Orange} refer to the results of DINO and $\mathcal{F}$-DINO, respectively.}
    \label{fig: Joint distribution plot}
\end{figure}

\begin{figure*}
    \centering
    \includegraphics[width=1\linewidth]{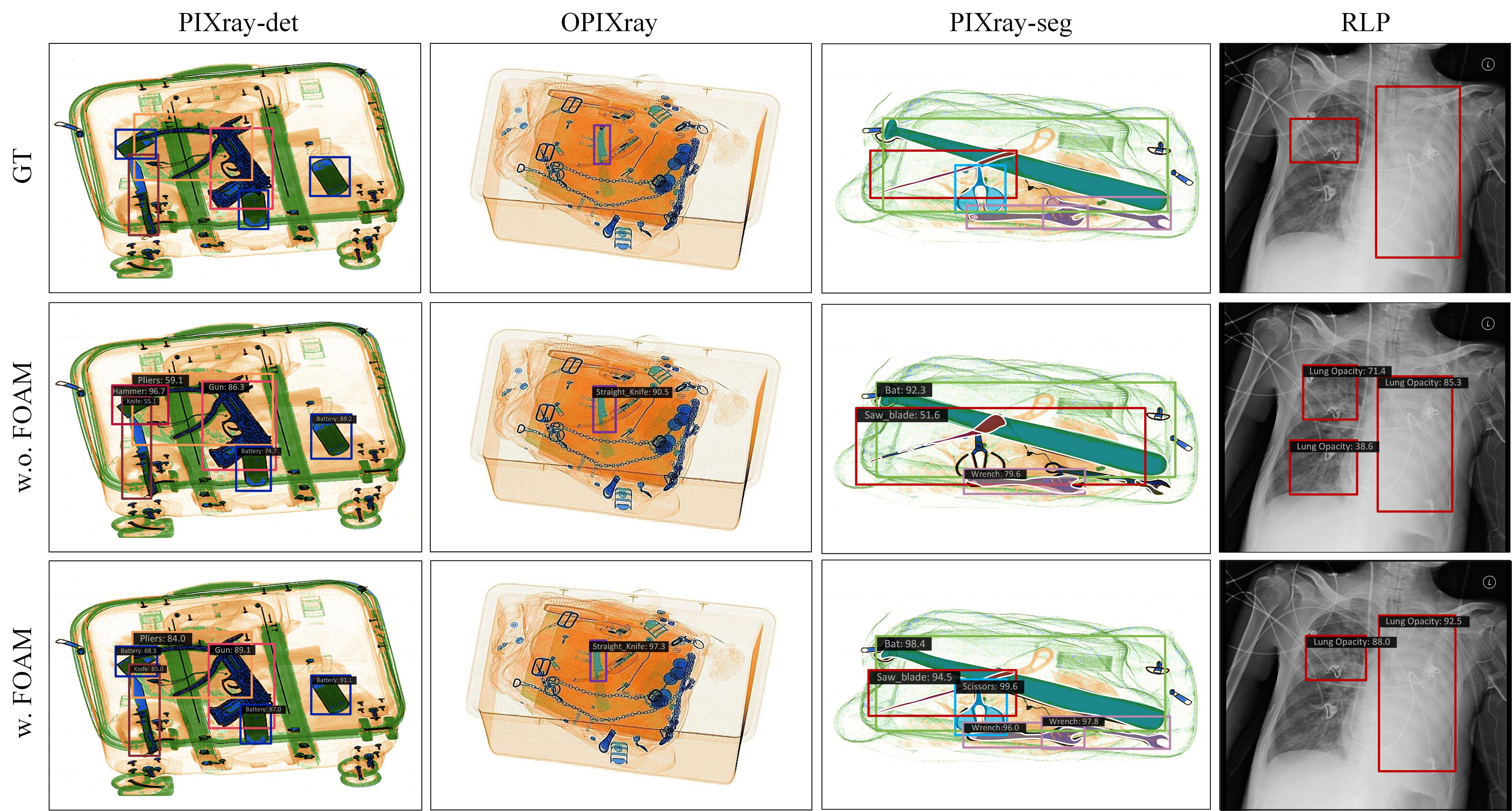}
    \caption{
    Prediction results on PIXray-det, OPIXray, RLP, PIXray-seg datasets. We select state-of-the-art models from four datasets across three tasks (row two) and apply FOAM to enhance their performance (row three). These models include DINO (Swin-L) on PIXray-det and OPIXray, Mask-R-CNN on PIXray-seg, and Cascade-R-CNN on RLP. In comparison, the models enhanced by FOAM achieve superior predictive accuracy.
    }
    \label{fig:visualization of perception results}
\end{figure*}
\subsubsection{Feature maps}\label{sec:Feature maps}
~\cref{fig:attention map visualization3} compares the feature maps of the DINO, DINO+FSTB, and DINO+FSTB+HDC ($\mathcal{F}$-DINO) models, all of which use the Swin-L backbone network. In row (a), the feature map generated by DINO captures incomplete ``Wrench'' features, with low response values and unclear contours. In contrast, the DINO+FSTB model demonstrates a higher feature response to the ``Wrench''. Furthermore, the feature map of $\mathcal{F}$-DINO shows greater contrast between the foreground and background, with clearer contours, indicating the positive guidance effect of the Hierarchical De-Corrupting (HDC) mechanism in enhancing FSTB’s ability to perceive features in the presence of overlap. Similarly, in row (b), the feature map of DINO fails to focus more on the prohibited items, such as the ``Battery'' and ``Screwdriver'', compared to the background. However, in the DINO+FSTB and $\mathcal{F}$-DINO feature maps, this focus gradually improves.
Finally, in row (c), DINO+FSTB demonstrates better attention to the heavily overlapping ``Gun'' and ``Hammer'' features compared to the baseline model. Additionally, $\mathcal{F}$-DINO, building upon this, further enhances attention to the ``Saw'' while reducing the focus on background objects.

Overall, the frequency domain features introduced by FSTB help the backbone network to more comprehensively perceive foreground features. 
The HDC mechanism essentially further directs FSTB to prioritize foreground feature attention while suppressing background feature focus, manifested as an enhancement in the perception of foreground contour, thereby improving the model's resistance to feature overlap.

\subsubsection{Sampling points and reference points}
We take DINO as a representative of the Deformable-DETR series models to explore how the reference point and sampling points of the final decoder layer, directly related to the final detection results, change in response to specific images under the influence of the proposed FOAM. As shown in~\cref{fig:sampling points}, column (a) indicates that the sampling points of $\mathcal{F}$-DINO are more concentrated on the prohibited item ``Fireworks'', while in the baseline model DINO, the high-confidence sampling points focus more on background features.
Further, in column (b), when faced with the weakly featured prohibited item ``Lighter'', severely disrupted by the background, the reference point of DINO accurately locates the target. However, the sampling points are not sufficiently concentrated, capturing a large amount of background features, which leads to the misclassification of the ``Lighter'' as ``Fireworks''. In contrast, $\mathcal{F}$-DINO's sampling points are not only more focused on the ``Lighter'' itself but also exhibit higher sampling confidence. This means that the ``Lighter'' features contribute more significantly to the model's final decision, resulting in a correct classification with high confidence and accurate localization~\cite{Deformable-DETR}.
In column (c), when faced with the overlapping and closely positioned ``Gun'' and ``Knife'', DINO's reference point and sampling points tend to focus on both the ``Knife'' and ``Gun'', leading to an incorrect detection result of ``Battery''. In contrast, reference point and sampling points of $\mathcal{F}$-DINO are focused on the ``Gun'' itself, resulting in the correct detection of the ``Gun''.

Overall, the frequency domain features supplemented by the FOAM backbone help the decoder focus on and extract foreground features from overlapping scenes, leading to more accurate predictions.
\subsubsection{The scatter diagram of prediction results}
To assess the effectiveness of the FOAM mechanism for prediction results, we visualize the IoU and classification scores of DINO and $\mathcal{F}$-DINO (DINO with our FOAM) predictions on the PIXray-det dataset, as shown in~\cref{fig: Joint distribution plot}. We plot a scatter diagram of prediction results with classification and IoU scores above 0.3, along with Kernel Density Estimation (KDE) curves. Blue and orange represent the results of DINO and $\mathcal{F}$-DINO, respectively.
The orange points are more concentrated, significantly shifted further to the right, and slightly moved upwards compared to the blue points, indicating that under FOAM, the model leverages both frequency and spatial domain cues to capture more classification semantic information, such as textures, as well as localization information, such as contours. This improvement enables the model to more effectively perceive and extract foreground information from overlapping features in complex scenes, thereby enhancing prediction accuracy for overlapping object perception.
\subsubsection{Prediction results}
~\cref{fig:visualization of perception results} illustrates the qualitative impact of FOAM on the prediction results of SOTA models across four datasets for three tasks.

\textbf{Prohibited Item Detection Task:} On the PIXray-det dataset, DINO (Swin-L) misclassifies the ``Battery'' in the top-left corner as a Hammer, and the localization result of the Pliers is disrupted by interference from the ``Gun''. Similarly, on the OPIXray dataset, the localization of the ``Straight Knife'' is compromised by background features. In contrast, under the influence of FOAM, $\mathcal{F}$-DINO demonstrates more accurate localization and classification for the prohibited item detection task, with higher confidence in its predictions.

\textbf{Prohibited Item Segmentation Task:} On the PIXray-seg dataset, Mask-R-CNN fails to detect the ``Wrench'' and ``Scissors'' and provides incomplete segmentation for the ``Saw''. In contrast, the instance segmentation results of $\mathcal{F}$-Mask-R-CNN are nearly identical to the ground truth.

\textbf{Pneumonia Detection Task:} On the RLP dataset, Cascade-R-CNN erroneously detects the background as ``Lung Opacity'', and its localization is imprecise. In comparison, $\mathcal{F}$-Cascade-R-CNN is less affected by interference from background elements~\cite{gambato2023chest} such as potential EKG leads, external tubes, artifacts, overlapping devices, bones, and healthy tissues, and more accurately detects the pathological boundaries of ``Lung Opacity''.

Overall, FOAM improves the accuracy of prediction results of SOTA models across multiple tasks and datasets, demonstrates its strong generalization ability, and enhances the perception of foreground features in overlapping scenes.
\section{Conclusion}
In this paper, we attempt to address the critical challenges of foreground-background feature coupling in overlapping object perception tasks.
Instead of adhering to mainstream spatial domain learning methods, we explore and leverage the advantages of frequency domain learning, designing a highly compatible joint perception approach in both the frequency and spatial domains, named the Frequency-Optimized Anti-Overlapping Framework (FOAM), which enhances the model's ability to perceive foreground textures and contours.
Extensive experimental results demonstrate that FOAM exhibits superior performance and a wide range of applications, significantly improving the accuracy of existing SOTA models across four datasets for at least three overlapping object perception tasks: Prohibited Item Detection, Prohibited Item Segmentation, and Pneumonia Detection.

\section*{Acknowledgments}
This work is supported by the National Natural Science Foundation of China under Grant U22A2063, 62173083, 62276186, and 62206043; the China Postdoctoral Science Foundation under No.2023M730517 and 2024T170114; the Liaoning Provincial "Selecting the Best Candidates by Opening Competition Mechanism" Science and Technology Program under Grant 2023JH1/10400045; the Fundamental Research Funds for the Central Universities under Grant N2424022; the Major Program of National Natural Science Foundation of China (71790614) and the 111 Project (B16009).



\bibliography{main}
\bibliographystyle{IEEEtran}

\vfill
\end{document}